\documentclass[10pt,twocolumn,letterpaper]{article}

\usepackage{cvpr}              

\usepackage{graphicx}
\usepackage{amsmath}
\usepackage{amssymb}
\usepackage{booktabs}

\usepackage[pagebackref,breaklinks,colorlinks]{hyperref}
\usepackage[accsupp]{axessibility} 
\usepackage{verbatim}
\usepackage{multirow, makecell}
\usepackage{lineno}
\usepackage{enumerate}
\usepackage{amsfonts}
\usepackage{gensymb}
\usepackage{bm}
\usepackage{bbding}
\usepackage{comment}
\usepackage{color}
\usepackage{diagbox}
\usepackage{tabularx}
\usepackage{rotating}
\newcolumntype{L}[1]{>{\raggedright\let\newline\\\arraybackslash\hspace{0pt}}m{#1}}
\newcolumntype{C}[1]{>{\centering\arraybackslash}m{#1}}
\newcolumntype{R}[1]{>{\raggedleft\let\newline\\\arraybackslash\hspace{0pt}}m{#1}}
\usepackage{xmpmulti}
\usepackage{arydshln}
\usepackage[dvipsnames]{xcolor}

\newlength\savewidth\newcommand\shline{\noalign{\global\savewidth\arrayrulewidth
  \global\arrayrulewidth 1pt}\hline\noalign{\global\arrayrulewidth\savewidth}}
\newcommand{\tablestyle}[2]{\setlength{\tabcolsep}{#1}\renewcommand{\arraystretch}{#2}\centering\footnotesize}

\usepackage[capitalize]{cleveref}
\crefname{section}{Sec.}{Secs.}
\Crefname{section}{Section}{Sections}
\Crefname{table}{Table}{Tables}
\crefname{table}{Tab.}{Tabs.}

\def\cvprPaperID{4495} 
\def\confName{CVPR}
\def\confYear{2022}

\begin{document}

\title{Hybrid Relation Guided Set Matching for Few-shot Action Recognition}

\author{
    Xiang Wang$^{1}$
     \hspace{0.2cm} 
    Shiwei Zhang$^{2*}$
     \hspace{0.2cm} 
    Zhiwu Qing$^{1}$
     \hspace{0.2cm} 
    Mingqian Tang$^2$
     \hspace{0.2cm}
    Zhengrong Zuo$^1$\\
     \hspace{0.2cm} 
    Changxin Gao$^1$ 
     \hspace{0.2cm}
     Rong Jin$^2$
    \hspace{0.2cm}
     Nong Sang$^{1*}$ \vspace{1mm} \\ 
    $^1$Key Laboratory of Image Processing and Intelligent Control,\\ School of Artificial Intelligence and Automation, Huazhong University of Science and Technology\\
    \hspace{0.3cm} $^2$Alibaba Group
    \\
{\tt\footnotesize \{wxiang,qzw,zhrzuo,cgao,nsang\}@hust.edu.cn, \{zhangjin.zsw,mingqian.tmq,jinrong.jr\}@alibaba-inc.com}
}

\maketitle
\let\thefootnote\relax\footnotetext{$*$ Corresponding authors.}
\begin{abstract}

Current few-shot action recognition methods reach impressive performance by learning discriminative features for each video via episodic training and designing various temporal alignment strategies.
%
%
Nevertheless, they are limited in that 
%
(a) learning individual features without considering the entire task may lose the most relevant information in the current episode,
and (b) these alignment strategies may fail in misaligned instances.
%
%
To overcome the two limitations, we propose a novel Hybrid Relation guided Set Matching (HyRSM) approach that incorporates two key components: hybrid relation module and set matching metric.
The purpose of the hybrid relation module is to learn task-specific embeddings by fully exploiting associated relations within and cross videos in an episode.
%
%
%
Built upon the task-specific features, we reformulate distance measure between query and support videos as a set matching problem and further design a bidirectional Mean Hausdorff Metric to improve the resilience to misaligned instances.
%
%
By this means, the proposed HyRSM can be highly informative and flexible to predict query categories under the few-shot settings.
We evaluate HyRSM on six challenging benchmarks, and the experimental results show its superiority over the state-of-the-art methods by a convincing margin.
Project page: \url{https://hyrsm-cvpr2022.github.io/}.
%
%
\end{abstract}

\section{Introduction}
\label{sec:intro}
%

%
Action recognition has been witnessing remarkable progress with the evolution of large-scale datasets~\cite{Kinetics,SSV2,EPIC-100} and video models~\cite{TSN,TSM,Slowfast}.
However, this success heavily relies on a large amount of manually labeled examples, which are labor-intensive and time-consuming to collect.
%
It actually limits further applications of this task.
Few-shot action recognition is promising in reducing manual annotations and thus has attracted much attention recently~\cite{CMN,ARN-ECCV}. It aims at learning to classify unseen action classes with extremely few annotated examples. 
%
%
%
%
\begin{figure}[t]  
\centering
\includegraphics[width=0.45\textwidth]{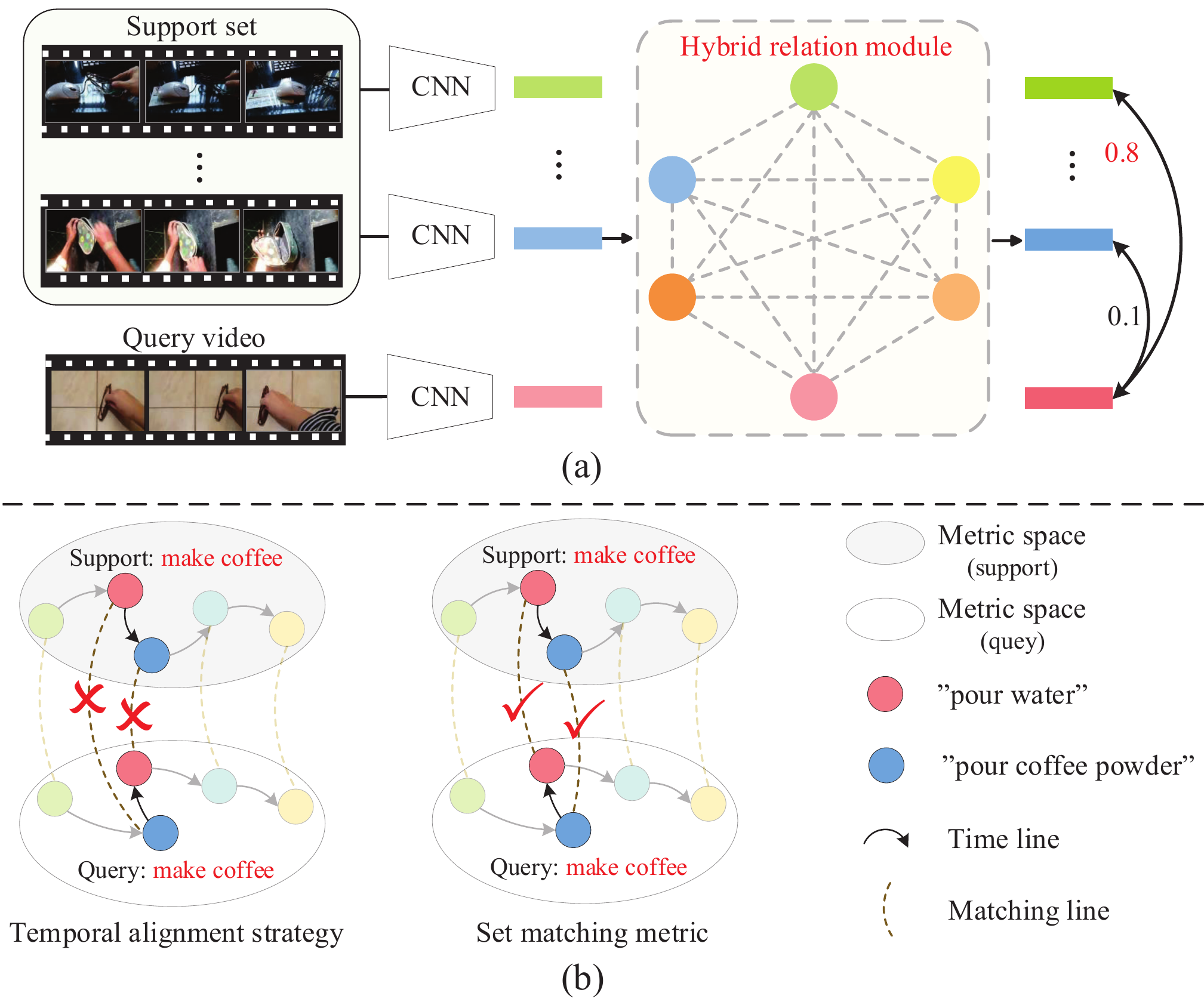}
\vspace{-4mm}
\caption{
%
%
(a) The proposed hybrid relation module.
We enhance video representations by extracting relevant discriminative patterns cross videos in an episode, which can   adaptively learn task-specific embeddings.
(b) Example of \emph{make coffee}, the current temporal alignment metrics tend to be strict, resulting in an incorrect match on misaligned videos.
In contrast, the proposed set matching metric is more flexible in finding the best correspondences.
}
\label{fig:Motivation}
\vspace{-5mm}
\end{figure}
%
%
%
%
%

%

To address the few-shot action recognition problem, current attempts~\cite{CMN,OTAM,TRX,ITANet} mainly adopt a metric-based meta-learning framework~\cite{prototypical} for its simplicity and effectiveness. 
It first learns a deep embedding space and then designs an explicit or implicit alignment metric to calculate the distances between the query (test) videos and support (reference) videos for classification in an episodic task.
For instance, Ordered Temporal Alignment Module (OTAM)~\cite{OTAM} extracts features for each video independently and tries to find potential query-support frame pairs only along the ordered temporal alignment path in this feature space.
Despite remarkable performance has been reached, these methods still suffer from two drawbacks.
First, discriminative interactive clues cross videos in an episode are ignored when each video is considered independently during representation learning.
As a result, these methods actually assume the learned representations are equally effective on different episodic tasks and maintain a fixed set of video features for all test-time tasks, \ie, task-agnostic, which hence might overlook the most discriminative dimensions for the current task.
Existing work also shows that the task-agnostic methods tend to suffer inferior generalization in other fields, such as image recognition~\cite{Finding_task-relevant, TapNet}, NLP~\cite{NLP_task-specific-1, NLP_task-specific-2}, and information retrieval~\cite{Retrieval_task-specific-1}.
Second, actions are usually complicated and involve many subactions with different orders and offsets, which may cause the failure of existing temporal alignment metrics.
For example, as shown in Figure~\ref{fig:Motivation}(b), to \emph{make coffee}, you can \emph{pour water} before \emph{pour coffee powder}, or in a reverse order, hence it is hard for recent temporal alignment strategies to find the right correspondences.
Thus a more flexible metric is required to cope with the misalignment.
%

%
%
%
Inspired by the above observations, we thus propose a novel Hybrid Relation guided Set Matching (HyRSM) algorithm that consists of a hybrid relation module and a set matching metric.
In the hybrid relation module, we argue that the considerable relevant relations within and cross videos are beneficial to generate a set of customized features that are discriminative for a given task.
%
To this end, we first apply an intra-relation function to strengthen structural patterns  within a video via modeling long-range temporal dependencies.
Then an inter-relation function operates on different videos to extract rich semantic information to reinforce the features which are more relevant to query predictions, as shown in Figure~\ref{fig:Motivation}(a).
By this means, we can learn task-specific embeddings for the few-shot task.
%
On top of the hybrid relation module, we design a novel bidirectional Mean Hausdorff Metric to calculate the distances between query and support videos from the set matching perspective.
%
Concretely, we treat each video as a set of frames and alleviate the strictly ordered constraints to acquire better query-support correspondences, as shown in Figure~\ref{fig:Motivation}(b).
In this way,  by combining the two components, the proposed HyRSM can sufficiently integrate semantically relational representations within the entire task and provide flexible video matching in an end-to-end manner.
We evaluate the proposed HyRSM on six challenging benchmarks and achieve remarkable improvements again current state-of-the-art methods.

Summarily, we make the following three contributions:
 1) We propose a novel hybrid relation module to capture the intra- and inter-relations inside the episodic task, yielding task-specific representations for different tasks.
 2) We further reformulate the query-support video pair distance metric as a set matching problem and develop a bidirectional Mean Hausdorff Metric, which can be robust to complex actions.
 3) We conduct extensive experiments on six challenging datasets to verify that the proposed HyRSM achieves superior performance over the state-of-the-art methods.
%
%

%
\section{Related Work}
\label{sec:related}
The work related to this paper includes: few-shot image classification, set matching, and few-shot action recognition. In this section, we will briefly review them separately. 

\vspace{+1pt}
\noindent \textbf{Few-shot Image Classification. } 
Recently, the research of few-shot learning~\cite{few-shot_feifei} has proceeded roughly along with the following directions: data augmentation, optimization-based, and metric-based.
Data augmentation is an intuitive method to increase the number of training samples and improve the diversity of data. 
Mainstream strategies include spatial deformation~\cite{augmentation_1,augmentation_2} and semantic feature augmentation~\cite{augmentation_feature_1,augmentation_feature_2}. 
Optimization-based methods learn a meta-learner model that can quickly adopt to a new task given a few training examples.
These algorithms include the LSTM-based meta-learner~\cite{meta-learner_1}, learning efficient model initialization~\cite{MAML}, and learning stochastic gradient descent optimizer~\cite{meta-learner_3}.
Metric-based methods attempt to address the few-shot classification problem by "learning to compare". 
This family of approaches aims to learn a feature space and compare query and support images through Euclidean distance~\cite{prototypical,TapNet}, cosine similarity~\cite{MatchNet,ye2020few}, or learnable non-linear metric~\cite{RelationNet,Cross_attention,Finding_task-relevant}.
Our work is more closely related to the metric-based methods~\cite{Finding_task-relevant,TapNet} that share the same spirit of learning task-specific features, whereas we focus on solving the more challenging few-shot action recognition task with diverse spatio-temporal dependencies.
In addition, we will further point out the differences and conduct performance comparisons in the supplementary materials.
%

\vspace{+1pt}
\noindent \textbf{Set Matching. }
The objective of set matching is to accurately measure the similarity of two sets, which have received much attention over the years.
Set matching techniques can be used to efficiently process complex data structures~\cite{setmatching_graph,setmatching_cluster,setmatching_graph_2} and has been applied in many computer vision fields, including face recognition~\cite{setmatching_face_1,setmatching_face_2,setmatching_face_3}, object matching~\cite{setmatching_object,Hausdorff_image_4}, \etc.
Among them, Hausdorff distance is an important alternative to handle set matching problems.
Hausdorff distance and its variants have been widely used in the field of image matching and achieved remarkable results~\cite{Hausdorff_image_1,Hausdorff_image_2,Hausdorff_image_3,Hausdorff_image_4,Hausdorff_image_5,Hausdorff_image_6}.
Inspired by these great successes, we introduce set matching into the few-shot action recognition field for the first time.
%
%
%
%
\begin{figure*}[t]  
\centering
\includegraphics[width=0.82\textwidth]{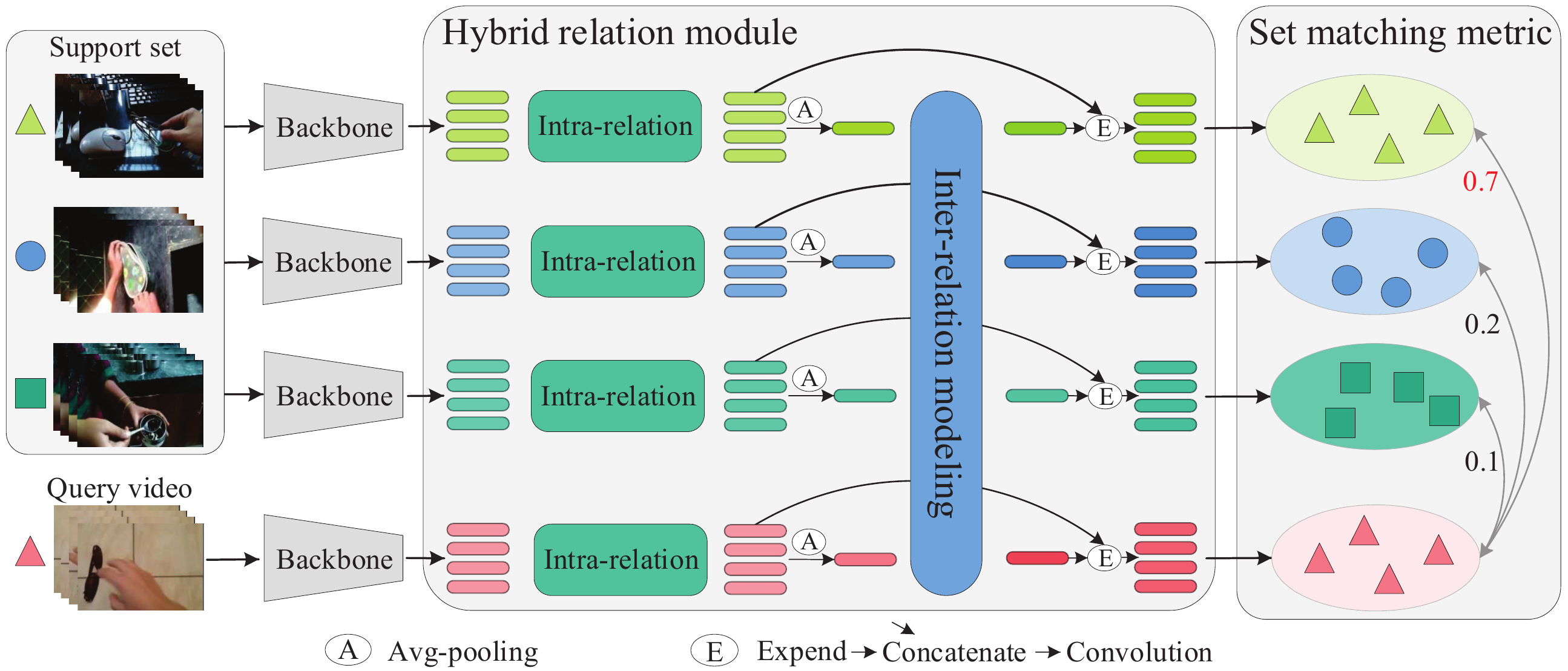}
\vspace{-3mm}
\caption{Schematic illustration of the proposed Hybrid Relation guided Set Matching (HyRSM) approach on a 3-way 1-shot problem. 
Given an episode of video data, a feature embedding network is first employed to extract their feature vectors.
A hybrid relation module is then followed to integrate rich information within each video and cross videos with intra-relation and inter-relation functions.
%
%
Finally, the task-specific features are fed forward into a set matching metric for matching score prediction.
%
Best viewed in color.
}
\label{fig:Network}
\vspace{-5mm}
\end{figure*}
%
%
%
%
%

\vspace{+1pt}
\noindent \textbf{Few-shot Action Recognition. }
The difference between few-shot action recognition and the previous few-shot learning approaches is that it deals with more complex higher-dimensional video data instead of two-dimensional images.
The existing methods mainly focus on metric-based learning.
OSS-Metric Learning~\cite{OSS-metric} adopts OSS-Metric of video pairs to match videos.
%
TARN~\cite{TARN} learns an attention-based deep-distance measure from an attribute to a class center for zero-shot and few-shot action recognition.
%
CMN~\cite{CMN} utilizes a multi-saliency embedding algorithm to encode video representations.
%
AMeFu-Net~\cite{few-shot-depth} uses depth information to assist learning.
OTAM~\cite{OTAM} preserves the frame ordering in video data and estimates distances with ordered temporal alignment.
ARN~\cite{ARN-ECCV} introduces a self-supervised permutation invariant strategy. 
ITANet~\cite{ITANet} proposes a frame-wise implicit temporal alignment strategy to achieve accurate and robust video matching.
TRX~\cite{TRX} matches actions by matching plentiful tuples of different sub-sequences.
Note that most above approaches focus on learning video embedding independently.
%
Unlike these previous methods, our HyRSM improves the transferability of embedding by learning intra- and inter-relational patterns that can better generalize to unseen classes.

\section{Method} 
\label{sec:method}
In this section, we first formulate the definition of the few-shot action recognition task.
Then we present our Hybrid  Relation  guided  Set  Matching  (HyRSM) method.

\subsection{Problem formulation} 
\label{subsec:definition}
%
The objective of few-shot action recognition is to learn a model that can generalize well to new classes with only a few labeled video samples.
To make training more faithful to the test environment, we adopt the episodic training manner~\cite{MatchNet} for few-shot adaptation as previous work~\cite{MatchNet,OTAM,TRX,ITANet}.
In each episodic task, there are two sets, \emph{i.e.}, a support set $S$ and a query set $Q$.
The support set $S$ contains $N \times K$ samples from $N$ different action classes, and each class contains $K$ support videos, termed the $N$-way $K$-shot problem.
The goal is to classify the query videos in $Q$ into $N$ classes with these support videos.

\subsection{HyRSM}

\textbf{Pipeline.}
The overall architecture of HyRSM is illustrated in Figure~\ref{fig:Network}. 
For each input video sequence, we first divide it into $T$ segments and extract a snippet from each segment, as in previous methods~\cite{TSN,OTAM}.
This way, in an episodic task, the support set can be denoted as $S=\{s_{1}, s_{2}, ..., s_{N \times K}\}$, where $s_{i} = \{s_{i}^{1}, s_{i}^{2}, ...,s_{i}^{T}\}$.
For simplicity and convenience, we discuss the process of the $N$-way $1$-shot problem, \ie, $K=1$, and consider that the query set $Q$ contains a single video $q$.
Then we apply an embedding model to extract the feature representations for each video sequence and obtain the support features $F_s=\{f_{s_1}, f_{s_2},...,f_{s_{N}}\}$ and the query feature $f_q$, where $f_{s_i}=\{f_{i}^{1}, f_{i}^{2}, ...,f_{i}^{T}\}$ and $f_q=\{f_{q}^{1}, f_{q}^{2}, ...,f_{q}^{T}\}$.
After that, we input $F_s$ and $f_q$ to the hybrid relation module to learn task-specific features, resulting in $\tilde{F}_s$ and $\tilde{f}_q$.
Finally, the enhanced representations  $\tilde{F}_s$ and $\tilde{f}_q$ are fed into the set matching metric to generate matching scores.
Based on the output scores, we can train or test the total framework.

\textbf{Hybrid relation module. }
Given the features $F_s$ and $f_q$ output by the embedding network, current approaches, \eg, OTAM~\cite{OTAM}, directly apply a classifier $\mathcal{C}$ in this feature space.
They can be formulated as:
\begin{equation}
y_{i} = \mathcal{C}(f_{s_i}, f_q)
\label{eq:prb_baseline}
\end{equation}
where $y_i$ is the matching score between $f_{s_{i}}$ and $f_q$.
During training, $y_i=1$ if they belong to the same class, otherwise $y_i=0$.
In the testing phase, $y_i$ can be adopted to predict the query label.
From the perspective of probability theory, it makes decisions based on the priors $f_{s_i}$ and $f_q$:
%
\begin{equation}
y_i=\mathcal{P}((f_{s_i}, f_q)|f_{s_i}, f_q)
\label{eq:prb_baseline_pro}
\end{equation}
%
which is a typical task-agnostic method.
%
%
However, the task-agnostic embedding is often vulnerable to overfit irrelevant representations~\cite{Cross_attention, Finding_task-relevant} and may fail to transfer to unseen classes not yet observed in the training stage.
%

%
%
Unlike the previous methods, we propose to learn task-specific features for each target task.
To achieve this goal, we introduce a hybrid relation module to generate task-specific features by capturing rich information from different videos in an episode.
Specifically,  we elaborately design the hybrid relation module $\mathcal{H}$ in the following form:
\begin{equation}
\tilde{f}_i=\mathcal{H}(f_i, \mathcal{G}); f_i \in  [F_s, f_q], \mathcal{G} = [F_s, f_q]
\end{equation}
That is, we improve the feature $f_i$ by aggregating semantic information cross video representations, \ie, $\mathcal{G}$, in an episodic task, allowing the obtained task-specific feature $\tilde{f}_i$ to be more discriminative than the isolated feature. 
%
For efficiency, we further decompose hybrid relation module into two parts:
intra-relation function $\mathcal{H}_a$ and inter-relation function $\mathcal{H}_e$.
%
%

%
The intra-relation function aims to strengthen structural patterns within a video by capturing long-range temporal dependencies.
We express this process as:
\begin{equation}
    f_i^a = \mathcal{H}_{a}(f_i)
\end{equation}
here $f_i^a \in \mathcal{R}^{T\times C}$ is the output of $f_i$ through the intra-relation function and has the same shape as $f_i$.
Note that the intra-relation function has many alternative implements, including multi-head self-attention (MSA), Transformer~\cite{Transformer}, Bi-LSTM~\cite{BILSTM}, Bi-GRU~\cite{GRU}, \etc, which is incredibly flexible and can be any one of them.

Based on the features generated by the intra-relation function, an inter-relation function is deployed to semantically enhance the features cross different videos:
%
%
\begin{equation}
f_i^e = \mathcal{H}_i^e(f_i^a, \mathcal{G}^a) \ 
        = \sum_j^{|\mathcal{G}^a|}(\kappa(\psi(f_i^a), \psi(f_j^a)) * \psi(f_j^a))
\end{equation}
where $\mathcal{G}^a = [F_s^a, f_q^a]$, $\psi(\cdot)$ is a global average pooling layer, and $\kappa(f_i^a, f_j^a)$ is a learnable function that calculates the semantic correlation between $f_i^a$ and $f_j^a$.
The potential logic is that if the correlation score between $f_i^a$ and $f_j^a$ is high, \ie, $\kappa(f_i^a, f_j^a)$, it means they tend to have the same semantic content, hence we can borrow more information from $f_j^a$ to elevate the representation $f_i^a$, and vice versa.
In the same way, if the score $\kappa(f_i^a, f_i^a)$ is less than $1$, it indicates that some irrelevant information in $f_i^a$ should be suppressed.
%

%
In this way, we can improve the feature discrimination by taking full advantage of the limited samples in each episodic task. 
%
%
The inter-relation function also has similar implements with the intra-relation function but with a different target.
After the inter-relation function, we employ an Expend-Concatenate-Convolution operation to aggregate information, as shown in Figure~\ref{fig:Network}, where the output feature $\tilde{f_{i}}$ has the same shape as $f_i^e$.
In the form of prior, our method can be formulated as:
\begin{equation}
y_i=\mathcal{P}((\tilde{f}_{s_i}, \tilde{f_q})|\mathcal{H}(f_{s_i}, \mathcal{G}), \mathcal{H}(f_q, \mathcal{G})); \mathcal{G} = [F_s, f_q]
\label{eq:task_speci_prob}
\end{equation}
Intuitively, compared with Equation~\ref{eq:prb_baseline_pro}, it can be conducive to making better decisions because more priors are provided.
%
In particular, the hybrid relation module is a plug-and-play unit.
In the experiment, we will fully explore different configurations of the hybrid relation module and further investigate its insertablility.


\textbf{Set matching metric. }
Given the relation-enhanced features $\tilde{F_{s}}$ and $\tilde{f_{q}}$, we present a novel metric to enable efficient and flexible matching.
%
%
%
In this metric, we treat each video as a set of $T$ frames and reformulate distance measurement between videos as a set matching problem, which is robust to complicated instances, whether they are aligned or not.
%
%
Specifically, we achieve this goal by modifying the Hausdorff distance, which is a typical set matching approach.
%
The standard Hausdorff distance $\mathcal{D}$ can be formulated as:
\begin{equation}
  \begin{split}
   d(\tilde{f_i},\tilde{f_q}) &= \max_{\tilde{f_i^a}\in \tilde{f_i}}(\min_{\tilde{f_q^b}\in \tilde{f_q}} \begin{Vmatrix} \tilde{f_i^a}-\tilde{f_q^b} \end{Vmatrix}) \\
   d(\tilde{f_q},\tilde{f_i}) &= \max_{\tilde{f_q^b}\in \tilde{f_q}}(\min_{\tilde{f_i^a}\in \tilde{f_i}} \begin{Vmatrix} \tilde{f_q^b}-\tilde{f_i^a} \end{Vmatrix}) \\
   \mathcal{D} &= \max(d(\tilde{f_i},\tilde{f_q}), d(\tilde{f_q},\tilde{f_i}))
  \end{split}
\end{equation}
where $\tilde{f_i} \in \mathcal{R}^{T \times C}$ contains $T$ frame features, and $\begin{Vmatrix} \cdot \end{Vmatrix}$ is a distance measurement function, which is the cosine distance in our method.
%

%
However, the previous methods~\cite{hausdorff-modified,hausdorff-modified-1,hausdorff-modified-2,Hausdorff_image_2} pointed out that Hausdorff distance can be easily affected by noisy examples, resulting in inaccurate measurements. 
%
Hence they employ a directed modified Hausdorff distance that robust to noise as follows:
\begin{equation}
    d_{m}(\tilde{f_i},\tilde{f_q}) = \frac{1}{N_i} \sum_{\tilde{f_i^a} \in \tilde{f_i}}(\min_{\tilde{f_q^b} \in \tilde{f_q}} 
    \begin{Vmatrix}
       \tilde{f_i^a}-\tilde{f_q^b}
   \end{Vmatrix}) 
\end{equation}
where $N_i$ is the length of $\tilde{f_i}$, and equal to $T$ in this paper.
%
Hausdorff distance and its variants achieve great success in image matching~\cite{Hausdorff_image_5,Hausdorff_image_2,Hausdorff_image_1} and face recognition~\cite{hausdorff-modified-1, Hausdorff_image_6}. We thus propose to introduce the set matching strategy into the few-shot action recognition field and further design a novel bidirectional Mean Hausdorff Metric (Bi-MHM):
%
\begin{equation}
\label{equ:bimhd}
 \begin{split}
    \mathcal{D}_b &= \frac{1}{N_i} \sum_{\tilde{f_i^a} \in \tilde{f_i}}(\min_{\tilde{f_q^b}\in \tilde{f_q}} \begin{Vmatrix}
       \tilde{f_i^a}-\tilde{f_q^b}
   \end{Vmatrix})  
   + \frac{1}{N_q} \sum_{\tilde{f_q^b} \in \tilde{f_q}}(\min_{\tilde{f_i^a} \in \tilde{f_i}} \begin{Vmatrix}
       \tilde{f_q^b} - \tilde{f_i^a}
   \end{Vmatrix})
 \end{split}
\end{equation}
%
%
where $N_i$ and $N_q$ are the lengths of the support feature $\tilde{f_i}$ and the query feature $\tilde{f_q}$ respectively.
\begin{table*}[ht]
\centering
\small
\tablestyle{6pt}{0.98}
\begin{tabular}{l|c|c|ccccc}
Method \hspace{4mm} &  \hspace{1mm} Reference \hspace{1mm} &  \hspace{2mm}  Dataset \hspace{2mm}  &  1-shot   &   2-shot   &   3-shot     &  4-shot     &  5-shot    \\ 
\shline
%
%
CMN++ \cite{CMN}  &  ECCV'18     &  \multirow{8}{*}{SSv2-Full}      & 34.4  & -  & -  & -  & 43.8  \\
TRN++ \cite{TRN-ECCV}  &  ECCV'18     &       & 38.6  & -  & -  & -  & 48.9  \\
OTAM \cite{OTAM}  &   CVPR'20    &        & 42.8  & 49.1  & 51.5  & 52.0  & 52.3  \\ 
TTAN \cite{TTAN}  &   ArXiv'21    &        & 46.3  & 52.5  & 57.3  & 59.3  & 60.4  \\
ITANet \cite{OTAM}  &   IJCAI'21    &        & \underline{49.2}  & \underline{55.5}  &\underline{59.1}  & 61.0  & 62.3  \\ 
TRX ($\Omega{=}\{1\}$) \cite{TRX}    & CVPR'21  &   & 38.8     & 49.7   & 54.4     & 58.0      & 60.6      \\ 
TRX ($\Omega{=}\{2,3\}$)\cite{TRX}  & CVPR'21   &        &  42.0   &  53.1   &  57.6   &   \underline{61.1}   &   \underline{64.6}   \\
\textbf{HyRSM}  & - &   & \hspace{3.7mm} \textbf{54.3}\color{blue}{ (+5.1)} \hspace{-4mm}  & \hspace{3.7mm} \textbf{62.2}\color{blue}{ (+6.7)} \hspace{-4mm} & \hspace{3.7mm} \textbf{65.1}\color{blue}{ (+6.0)} \hspace{-4mm}  &  \hspace{3.7mm} \textbf{67.9}\color{blue}{ (+6.8)} \hspace{-4mm} & \hspace{3.7mm} \textbf{69.0}\color{blue}{ (+4.4)} \hspace{-4mm} \\ 
\shline
MatchingNet \cite{MatchNet} & NeurIPS'16 &\multirow{11}{*}{Kinetics}      & 53.3  & 64.3  & 69.2  & 71.8  & 74.6  \\ 
MAML \cite{MAML}   & ICML'17   & & 54.2  & 65.5  & 70.0  & 72.1  & 75.3  \\ 
Plain CMN \cite{CMN}  & ECCV'18 & & 57.3  & 67.5  & 72.5  & 74.7  & 76.0  \\ 
CMN-J \cite{CMN-J}  & TPAMI'20 &   &  60.5   &    70.0   &    75.6   &    77.3   &    78.9   \\
TARN \cite{TARN}    & BMVC'19 & & 64.8  & -  & -  & -  & 78.5  \\ 
ARN \cite{ARN-ECCV} &  ECCV'20  &  & 63.7  & -  & -  & -  & 82.4  \\ 
OTAM \cite{OTAM}   & CVPR'20 &  & 73.0  & 75.9  & 78.7  & 81.9  & 85.8  \\ 
ITANet \cite{ITANet}   & IJCAI'21 &  & \underline{73.6}  & -  & -  & -  & 84.3  \\
TRX ($\Omega{=}\{1\}$) \cite{TRX}   & CVPR'21 &  & 63.6  & 75.4  & 80.1  & 82.4  & 85.2  \\ 
TRX ($\Omega{=}\{2,3\}$) \cite{TRX}  & CVPR'21 &   & 63.6  & \underline{76.2}  & \underline{81.8}  & \underline{83.4}  & \underline{85.9}  \\ 
\textbf{HyRSM}  & - &   & \hspace{3.7mm} \textbf{73.7}\color{blue}{ (+0.1)} \hspace{-4mm}  & \hspace{3.7mm} \textbf{80.0}\color{blue}{ (+3.8)} \hspace{-4mm} & \hspace{3.7mm} \textbf{83.5}\color{blue}{ (+1.7)} \hspace{-4mm}  &  \hspace{3.7mm} \textbf{84.6}\color{blue}{ (+1.2)} \hspace{-4mm} & \hspace{3.7mm} \textbf{86.1}\color{blue}{ (+0.2)} \hspace{-4mm} \\ 
\shline  
OTAM \cite{OTAM}  &   CVPR'20    &    \multirow{3}{*}{Epic-kitchens}     & \underline{46.0}  & 50.3  & \underline{53.9}  & 54.9  & 56.3  \\ 
TRX \cite{TRX}  & CVPR'21   &        &  43.4   &  \underline{50.6}   &  53.5   &   \underline{56.8}   &   \underline{58.9}   \\
\textbf{HyRSM}  & - &   & \hspace{3.7mm} \textbf{47.4}\color{blue}{ (+1.4)} \hspace{-4mm}  & \hspace{3.7mm} \textbf{52.9}\color{blue}{ (+2.3)} \hspace{-4mm} & \hspace{3.7mm} \textbf{56.4}\color{blue}{ (+2.5)} \hspace{-4mm}  &  \hspace{3.7mm} \textbf{58.8}\color{blue}{ (+2.0)} \hspace{-4mm} & \hspace{3.7mm} \textbf{59.8}\color{blue}{ (+0.9)} \hspace{-4mm} \\ 
%
%
\shline
ARN \cite{ARN-ECCV}  &   ECCV'20    &    \multirow{5}{*}{HMDB51}     & 45.5  & -  & -  & -  & 60.6  \\ 
OTAM \cite{OTAM}  &   CVPR'20    &         & 54.5  & \underline{63.5}  & 65.7  & 67.2  & 68.0  \\
TTAN \cite{TTAN}  &   ArXiv'21    &         & \underline{57.1}  & -  & -  & -  & 74.0  \\  
TRX \cite{TRX}  & CVPR'21   &        &  53.1   &  62.5   &  \underline{66.8}   &   \underline{70.2}   &   \underline{75.6}   \\
\textbf{HyRSM}  & - &   & \hspace{3.7mm} \textbf{60.3}\color{blue}{ (+3.2)} \hspace{-4mm}  & \hspace{3.7mm} \textbf{68.2}\color{blue}{ (+4.7)} \hspace{-4mm} & \hspace{3.7mm} \textbf{71.7}\color{blue}{ (+4.9)} \hspace{-4mm}  &  \hspace{3.7mm} \textbf{75.3}\color{blue}{ (+5.1)} \hspace{-4mm} & \hspace{3.7mm} \textbf{76.0}\color{blue}{ (+0.4)} \hspace{-4mm} \\ 
\end{tabular}
\vspace{-4mm}
\caption{Comparison to recent few-shot action recognition methods on the meta-testing set of SSv2-Full, Kinetics, Epic-kitchens and HMDB51. 
The experiments are conducted under the 5-way setting, and results are reported as the shot increases from 1 to 5.
"-" means the result is not available in published works, and the underline indicates the second best result.
}
\label{tab:compare_SOTA_1}
\vspace{-5mm}
\end{table*}

The proposed Bi-MHM is a symmetric function, and the two items are complementary to each other.
%
%
From Equation~\ref{equ:bimhd}, we can find that $\mathcal{D}_b$ can automatically find the best correspondencies between two videos, \eg, $\tilde{f_i}$ and $\tilde{f_q}$.
Note that our Bi-MHM is a non-parametric classifier and does not involve numerous non-parallel calculations, which helps to improve computing efficiency and transfer ability compared to the previous complex alignment classifiers~\cite{OTAM,TRX}.
%
Moreover, the hybrid relation module and Bi-MHM can mutually reinforce each other, consolidating the correlation between two videos collectively. 
%
%
In the training phase, we take the negative distance for each class as logit.
Then we utilize the same cross-entropy loss as in~\cite{OTAM,TRX} and the regularization loss~\cite{few-shot-regular,few-shot-regular-2} to train the model.
The regularization loss refers to the cross-entropy loss on the real action classes, which is widely used to improve the training stability and generalization.
During inference, we select the support class closest to the query for classification.
%
%

%
%
%

\section{Experiments}
\label{sec:experiments}
The experiments are designed to answer the following
key questions: 
(1) Is HyRSM competitive to other state-of-the-art methods on challenging few-shot benchmarks? 
(2) What are the essential components and factors that make HyRSM work?
%
(3) Can the hybrid relation module be utilized as a simple plug-and-play component and bring benefits to existing methods?
(4) Does the proposed set matching metric have an advantage over other competitors?
%

\vspace{-1mm}
\subsection{Datasets and experimental setups}
\vspace{-1mm}
\noindent \textbf{Datasets.}
We evaluate our method on six few-shot datasets.
%
For the Kinetics~\cite{Kinetics}, SSv2-Full~\cite{SSV2}, and SSv2-Small~\cite{SSV2} datasets, we adopt the existing splits proposed by~\cite{OTAM,CMN,ITANet,TRX}, and each dataset consists of 64 and 24 classes as the meta-training and meta-testing set, respectively.
For UCF101~\cite{UCF101} and HMDB51~\cite{HMDB51}, we evaluate our method by using splits from~\cite{ARN-ECCV,TRX}.
%
In addition, we also use the Epic-kitchens~\cite{EPIC-100-2,EPIC-100} dataset to evaluate HyRSM. 
Please see the supplementary materials for more details.
%
%

%
%
\begin{table*}[ht]
\centering
\small
\tablestyle{6pt}{1.00}
\begin{tabular}{l|c|ccc|ccc}
\multicolumn{1}{l}{} &  \multicolumn{1}{l}{} & \multicolumn{3}{c|}{UCF101}  & \multicolumn{3}{c}{SSv2-Small} \\
\cmidrule(){3-5} 
\cmidrule(){6-8}
\hspace{-0mm} Method \hspace{2mm} &  \hspace{1mm} Reference \hspace{1mm} & \hspace{0mm}  1-shot \hspace{0mm}  &   3-shot & \hspace{3.5mm} 5-shot \hspace{3.5mm}  &  1-shot  & 3-shot & 5-shot \\

\shline
\hspace{-0mm} MatchingNet \cite{MatchNet} & NeurIPS'16       & -  & - & - & 31.3 & 39.8 &  45.5 \\ 
\hspace{-0mm} MAML \cite{MAML}   & ICML'17    & -  & -  & - & 30.9 & 38.6 & 41.9 \\ 
\hspace{-0mm} Plain CMN \cite{CMN}  & ECCV'18  & -  & - & - & 33.4 & 42.5 & 46.5 \\ 
\hspace{-0mm} CMN-J \cite{CMN-J}  & TPAMI'20    &  -   & -  &  - & 36.2 & 44.6 & 48.8  \\
\hspace{-0mm} ARN \cite{ARN-ECCV} &  ECCV'20   & 66.3  & -  & 83.1 & - & - & -  \\ 
\hspace{-0mm} OTAM \cite{OTAM}   & CVPR'20 &  79.9  & 87.0  & 88.9 & 36.4 & 45.9 & 48.0 \\ 
\hspace{-0mm} TTAN \cite{TTAN}   & ArXiv'21 &  \underline{80.9}  & -  & 93.2 & - & - & -  \\
\hspace{-0mm} ITANet \cite{ITANet}   & IJCAI'21   & - & - & - & \underline{39.8} & 49.4 &  53.7 \\
\hspace{-0mm} TRX \cite{TRX}  & CVPR'21 &    78.2  & \underline{92.4}  & \textbf{96.1} & 36.0 & \underline{51.9} & \textbf{59.1} \\ 
\hspace{-0mm} \textbf{HyRSM}  & - &   \hspace{3.7mm} \textbf{83.9}\color{blue}{ (+3.0)} \hspace{-4mm}  & \hspace{3.7mm} \textbf{93.0}\color{blue}{ (+0.6)} \hspace{-4mm} & \hspace{3.7mm} \underline{94.7}\color{red}{ (-1.4)} \hspace{-3mm} & \hspace{3.7mm} \textbf{40.6}\color{blue}{ (+0.8)} \hspace{-4mm} & \hspace{3.7mm} \textbf{52.3}\color{blue}{ (+0.4)} \hspace{-4mm} & \hspace{4mm} \underline{56.1}\color{red}{ (-3.0)} \hspace{-3mm} \\ 

\end{tabular}
\vspace{-3mm}
\caption{Results on 1-shot, 3-shot, and 5-shot few-shot classification on the UCF101 and SSv2-Small datasets. 
%
"-" means the result is not available in published works, and the underline indicates the second best result.
}
\vspace{-3mm}
\label{tab:compare_SOTA_2}
\end{table*}

\noindent \textbf{Implementation details.}
Following previous works~\cite{CMN,OTAM,TRX,ITANet}, we utilize ResNet-50~\cite{Resnet} as the backbone which
is initialized with ImageNet~\cite{imagenet} pre-trained weights. 
We sparsely and uniformly sample 8 (\ie, $T=8$) frames per video, as in previous methods~\cite{OTAM, ITANet}.
In the training phase, we also adopt basic data augmentation such as random cropping and color jitter,  
and we use Adam~\cite{adam} optimizer to train our model. 
For inference, we conduct few-shot action recognition evaluation on 10000 randomly sampled episodes from the meta-testing set and report the mean accuracy.
For many shot classification, \eg, 5-shot, we follow ProtoNet~\cite{prototypical} and calculate the mean features of support videos in each class as the prototypes, and classify the query videos according to their distances against the prototypes.

\subsection{Comparison with state-of-the-art}
We compare the performance of HyRSM with state-of-the-art methods in this section. 
As shown in Table~\ref{tab:compare_SOTA_1} and Table~\ref{tab:compare_SOTA_2}, our proposed HyRSM outperforms other methods significantly and achieves new state-of-the-art performance.
For instance, HyRSM improves the state-of-the-art performance from 49.2\% to 54.3\% under the 1-shot setting on SSv2-Full.
Specially, compared with the temporal alignment methods~\cite{OTAM,ITANet} and complex fusion methods~\cite{TTAN,TRX}, HyRSM consistently surpasses them under most different shots, which implies that our approach is considerably flexible and efficient.
%
%
%
Note that the SSv2-Full and SSv2-Small datasets tend to be motion-based and generally focus on temporal reasoning. 
While Kinetics and UCF101 are partly appearance-related datasets, and scene understanding is usually important.
%
Besides, Epic-kitchens and HMDB51 are relatively complicated and might involve diverse object interactions.
Excellent performance on these datasets reveals that our HyRSM has strong robustness and generalization for different scenes.
%
%
%
From Table~\ref{tab:compare_SOTA_2}, we observe that HyRSM outperforms current state-of-the-art methods on UCF101 and SSv2-Small under the 1-shot and 3-shot settings, which suggests that our HyRSM can learn rich and effective representations with extremely limited samples.
Of note, our HyRSM achieves 94.7\% and 56.1\% 5-shot performance on UCF101 and SSv2-Small, respectively, which is slightly behind TRX. 
We attribute this to TRX is an ensemble method specially designed for multiple shots.
%
%
%
%
\begin{figure}[t]  
\centering
\includegraphics[width=0.35\textwidth]{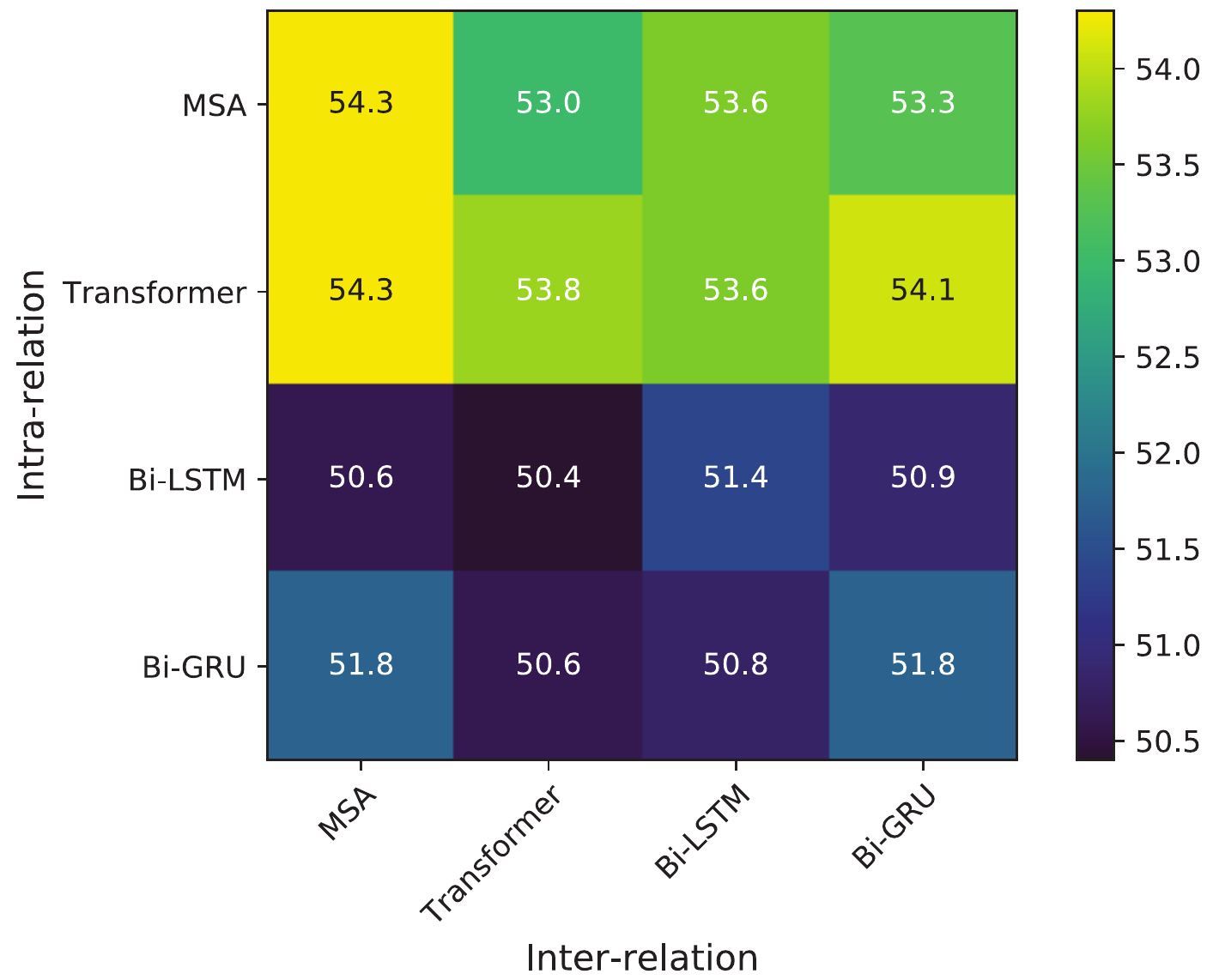}
\vspace{-3mm}
\caption{Comparison between different components in hybrid relation module on
5-way 1-shot few-shot action classification. Experiments are conducted on the SSv2-Full dataset.}
\label{fig:intra-intra}
\vspace{-4mm}
\end{figure}
%
%
%
%
\begin{table}[t]
\centering
\small
\tablestyle{6pt}{1.05}
\begin{tabular}{ccc|cc}
\hspace{-1mm} Intra-relation  &  Inter-relation & Bi-MHM &   1-shot  &  5-shot \\

\shline
\hspace{-1mm}  &  &  & 35.2  & 45.3 \\
\shline
\hspace{-1mm} \CheckmarkBold &  &  & 41.2  & 55.0 \\
\hspace{-1mm}  & \CheckmarkBold &  & 43.7  & 55.2 \\
\hspace{-1mm}  &  & \CheckmarkBold & 44.6  & 56.0 \\
\hspace{-1mm} \CheckmarkBold & \CheckmarkBold &  & 48.1  & 60.5 \\
\hspace{-1mm}  & \CheckmarkBold & \CheckmarkBold & 48.3  & 61.2 \\
\hspace{-1mm} \CheckmarkBold &  & \CheckmarkBold & 51.4  & 64.6 \\
\hspace{-1mm} \CheckmarkBold & \CheckmarkBold & \CheckmarkBold & \textbf{54.3}  & \textbf{69.0} 
\end{tabular}
\vspace{-2mm}
\caption{Ablation study under 5-way 1-shot and 5-way 5-shot settings on the SSv2-Full dataset.
}
\label{tab:ablation}
\vspace{-5mm}
\end{table}

%
%
%
\begin{table}[t]
\centering
\small
\tablestyle{6pt}{1.1}
\begin{tabular}{l|cc}
\hspace{-1mm} Method  &   1-shot  &  5-shot \\

\shline
\hspace{-1mm}  OTAM~\cite{OTAM}   & 42.8  & 52.3 \\
\hspace{-1mm} OTAM~\cite{OTAM}+ Intra-relation  & 48.9  & 60.4 \\
\hspace{-1mm}  OTAM~\cite{OTAM}+ Inter-relation   & 46.9  & 57.8 \\
\hspace{-1mm} OTAM~\cite{OTAM}+ Intra-relation + Inter-relation & 51.7  & 63.9 \\
\end{tabular}
\vspace{-2mm}
\caption{Generalization of hybrid relation module. We conduct experiments on SSv2-Full.
}
\label{tab:generalization}
\vspace{-5mm}
\end{table}

\subsection{Ablation study}
%
%
%
For ease of comparison, we use a baseline method ProtoNet~\cite{prototypical} that applies global-average pooling to backbone representations to obtain a prototype for each class.
\vspace{+3pt}
\noindent \textbf{Design choices of relation modeling. }
As shown in Figure~\ref{fig:intra-intra}, we vary the components in the hybrid relation module and systematically evaluate the effect of different variants. 
The experiments are performed on SSv2-Full under the 5-way 1-shot setting.
We can observe that different combinations have quite distinct properties, \eg, multi-head self-attention (MSA) and Transformer are more effective to model intra-class relations than Bi-LSTM and Bi-GRU. 
%
%
%
Nevertheless, compared with other recent methods~\cite{TRX,ITANet}, the performance of each combination can still be improved, which benefits from the effectiveness of structure design for learning task-specific features.
For simplicity, we adopt the same structure to model intra-relation and inter-relation, and we choose multi-head self-attention in the experiments.

\vspace{+3pt}
\noindent \textbf{Analysis of the proposed components. }
%
Table~\ref{tab:ablation} summarizes the effects of each module in HyRSM.
We take ProtoNet~\cite{prototypical} as our baseline method.
%
%
From the results, we observe that each component is highly effective.
%
In particular, compared to baseline, intra-relation modeling can respectively bring 6\% and 9.7\% performance gain on 1-shot and 5-shot, and inter-relation function boosts the performance by 8.5\% and 9.9\% on 1-shot and 5-shot.
In addition, the proposed set matching metric improves on 1-shot and 5-shot by 9.4\% and 10.7\%, respectively, which indicates the ability to find better corresponding frames in the video pair.
Moreover, stacking modules can further improve performance, indicating the complementarity between components.
%
%
\begin{figure}[t]  
\centering
\includegraphics[width=0.44\textwidth]{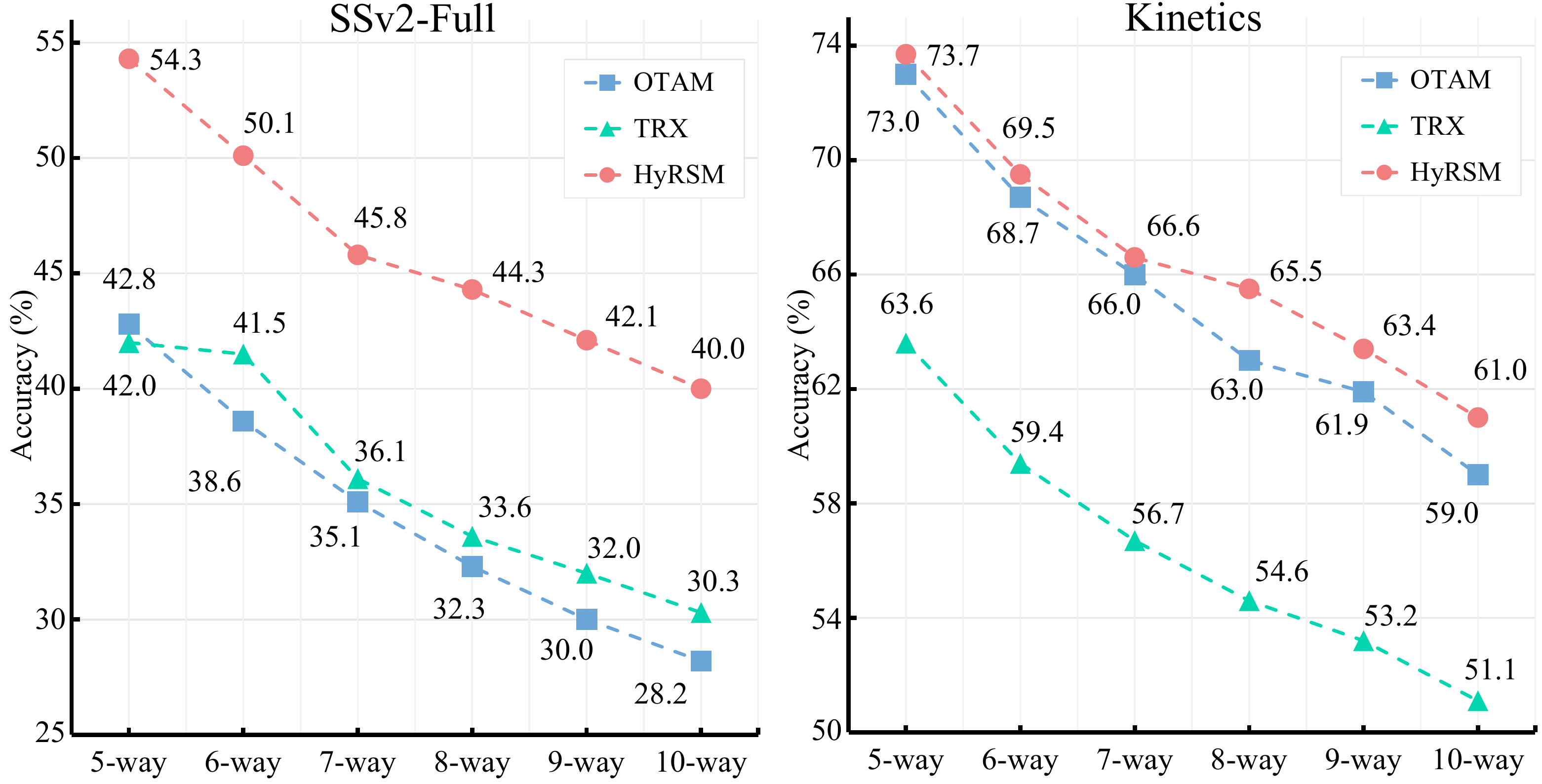}
\vspace{-3mm}
\caption{N-way 1-shot performance trends of our HyRSM and other state-of-the-art methods with different N on SSv2-Full.
The comparison results prove the superiority of our HyRSM}
\label{fig:Nway1shot}
\vspace{-3mm}
\end{figure}

%
%
%
%
%
\begin{figure}[t]  
\centering
\includegraphics[width=0.44\textwidth]{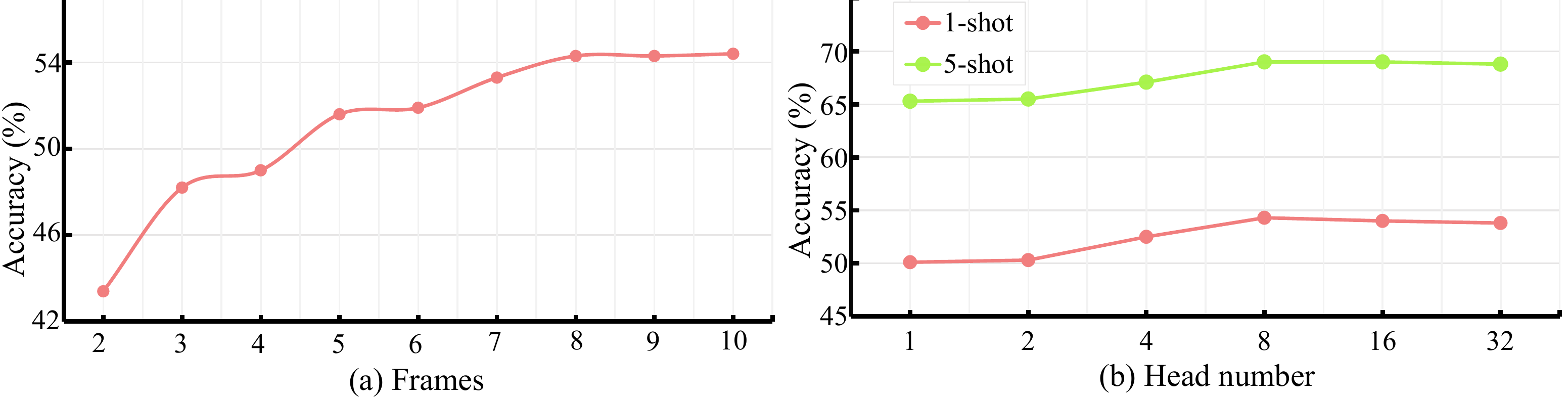}
\vspace{-3mm}
\caption{(a) Performance on SSv2-Full using a different number of frames under the 5-way 1-shot setting. (b) The effect of the number of heads on SSv2-Full.}
\label{fig:5-way_1-shot_mframes&head}
\vspace{-3mm}
\end{figure}
%
%

\vspace{+3pt}
\noindent \textbf{Pluggability of hybrid relation module. }
In Table~\ref{tab:generalization}, we experimentally show that the hybrid relation module generalizes well to other methods by inserting it into the recent OTAM~\cite{OTAM}.
%
In this study, OTAM with our hybrid relation module benefits from relational information and finally achieves 8.9\% and 11.6\% gains on 1-shot and 5-shot.
This fully evidences that mining the rich information among videos to learn task-specific features is especially valuable.

\vspace{+3pt}
\noindent \textbf{N-way few-shot classification. }
%
In the previous experiments, all of our comparative evaluation experiments were carried out under the 5-way setting. 
In order to further explore the influence of different N, in Figure~\ref{fig:Nway1shot}, we compare N-way (N $\ge$ 5) 1-shot results on SSv2-Full and Kinetics.
Results show that as N increases, the difficulty becomes higher, and the performance decreases.
Nevertheless, the performance of our HyRSM is still consistently ahead of the recent state-of-the-art OTAM~\cite{OTAM} and TRX~\cite{TRX}, which shows the feasibility of our method to boost performance by introducing rich relations among videos and the power of the set matching metric.

\vspace{+3pt}
\noindent \textbf{Varying the number of frames. }
To demonstrate the scalability of HyRSM, we also explore the impact of different video frame numbers on performance.
Of note, previous comparisons are performed under 8 frames of input.
Results in Figure~\ref{fig:5-way_1-shot_mframes&head}(a) show that as the number of frames increases, the performance improves. 
HyRSM gradually tends to be saturated when more than 8 frames.

\vspace{+3pt}
\noindent \textbf{Influence of head number. }
Previous analyses have shown that multi-head self-attention can focus on different patterns and is critical to capturing diverse features~\cite{multi-head-1}.
We investigate the effect of varying the number of heads in multi-head self-attention on performance in Figure~\ref{fig:5-way_1-shot_mframes&head}(b). 
Results indicate that the effect of multi-head is significant, and the performance starts to saturate beyond a particular point.
%
%
%
\begin{table}
\centering
\small
\tablestyle{6pt}{1.1}
\begin{tabular}{l|c|cc}
\hspace{-1mm} Metric \hspace{2mm} &  \hspace{0mm} Bi-direction \hspace{0mm} &   1-shot  &  5-shot \\

\shline
\hspace{-1mm} Diagonal & - &  38.3  & 48.7 \\
\hspace{-1mm} Plain DTW~\cite{DTW} & - &  39.6  & 49.0 \\
\hspace{-1mm} OTAM~\cite{OTAM} & \XSolidBrush &  39.3  & 47.7 \\
\hspace{-1mm} OTAM~\cite{OTAM} & \CheckmarkBold &  \underline{42.8}  & \underline{52.3} \\
\hspace{-1mm} \textbf{Bi-MHM (ours)} & \CheckmarkBold & \textbf{44.6} & \textbf{56.0} 
\end{tabular}
\vspace{-3mm}
\caption{Comparison with recent temporal alignment methods on the SSv2-Full dataset under the 5-way 1-shot and 5-way 5-shot settings. 
Diagonal means matching frame by frame.
}
\label{tab:matching-alignment}
\vspace{-1mm}
\end{table}

%
\begin{table}
\centering
\small
\tablestyle{6pt}{1.1}
\begin{tabular}{l|c|cc}
\hspace{-1mm} Metric \hspace{2mm} &  \hspace{0mm} Bi-direction \hspace{0mm} &   1-shot  &  5-shot \\

\shline
\hspace{-1mm} Hausdorff distance & \XSolidBrush & 32.4 & 38.2 \\
\hspace{-1mm} Hausdorff distance & \CheckmarkBold & 34.5 & 39.1 \\
\hspace{-1mm} Modified Hausdorff distance & \XSolidBrush & \underline{44.2} & \underline{50.0} \\
\hspace{-1mm} \textbf{Bi-MHM (ours)} & \CheckmarkBold & \textbf{44.6} & \textbf{56.0} 
\end{tabular}
\vspace{-3mm}
\caption{Comparison of different set matching strategies on the SSv2-Full dataset.
}
\label{tab:matching}
\vspace{-2mm}
\end{table}
%
%
\begin{figure}[t]  
\centering
\includegraphics[width=0.45\textwidth]{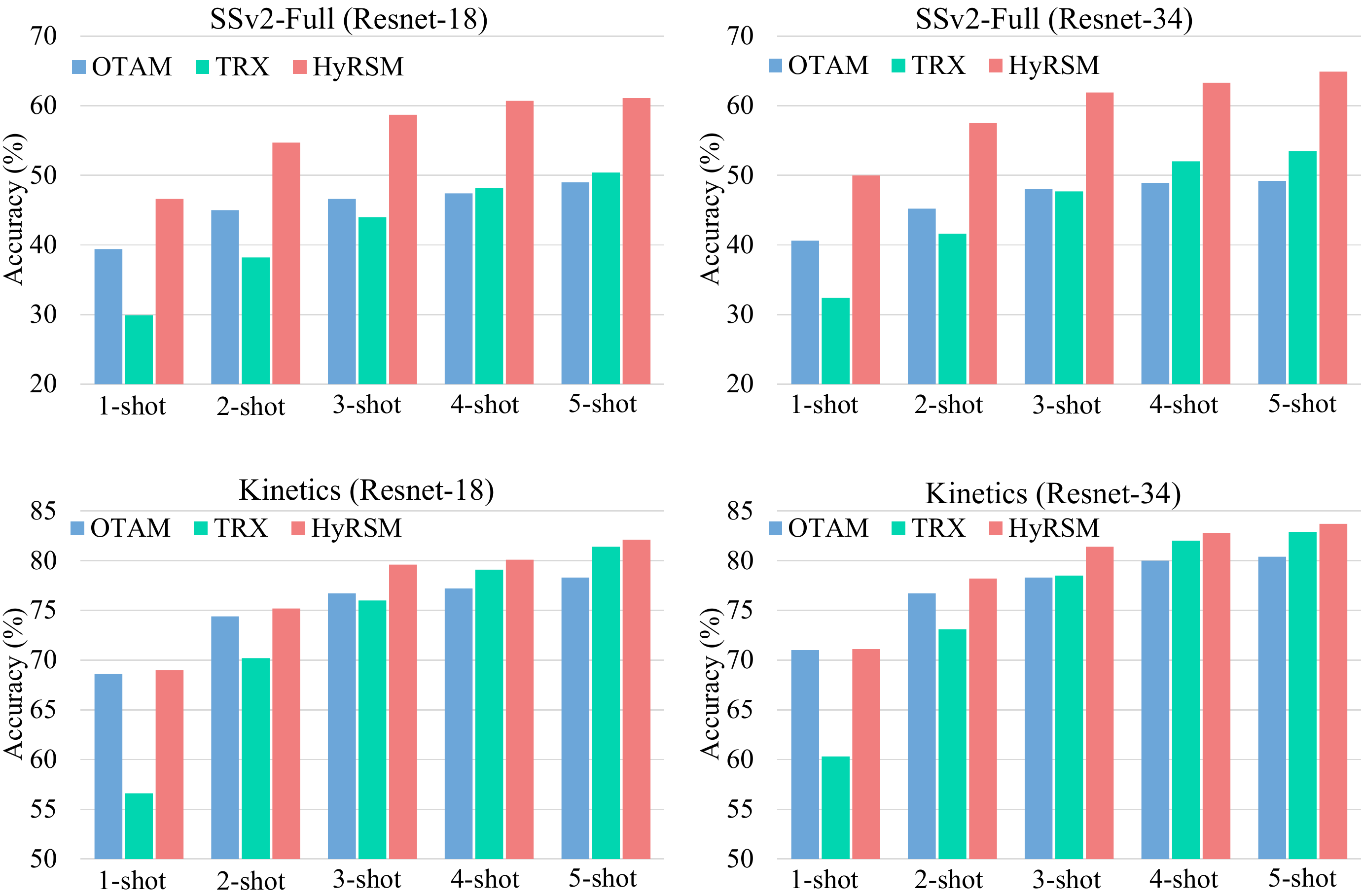}
\vspace{-2mm}
\caption{Comparison of the backbone with different depths.}
\label{fig:resnet18&resnet34}
\vspace{-3mm}
\end{figure}
%
%

\vspace{+3pt}
\noindent \textbf{Varying depth of the backbone. }
%
The previous methods all utilize ResNet-50 as backbone by default for a fair comparison, and the impact of backbone's depth on performance is still under-explored.
As presented in Figure~\ref{fig:resnet18&resnet34},  we attempt to answer this question by adopting ResNet-18 and ResNet-34 pre-trained on ImageNet as alternative backbones.
Results demonstrate that the deeper network clearly benefits from greater learning capacity and results in better performance.
%
%
In addition, we notice that our proposed HyRSM consistently outperforms the competitors (\ie, OTAM and TRX), which indicates that our HyRSM is a general framework.
%
%
%
%
%
\begin{figure}[t]  
\centering
\includegraphics[width=0.46\textwidth]{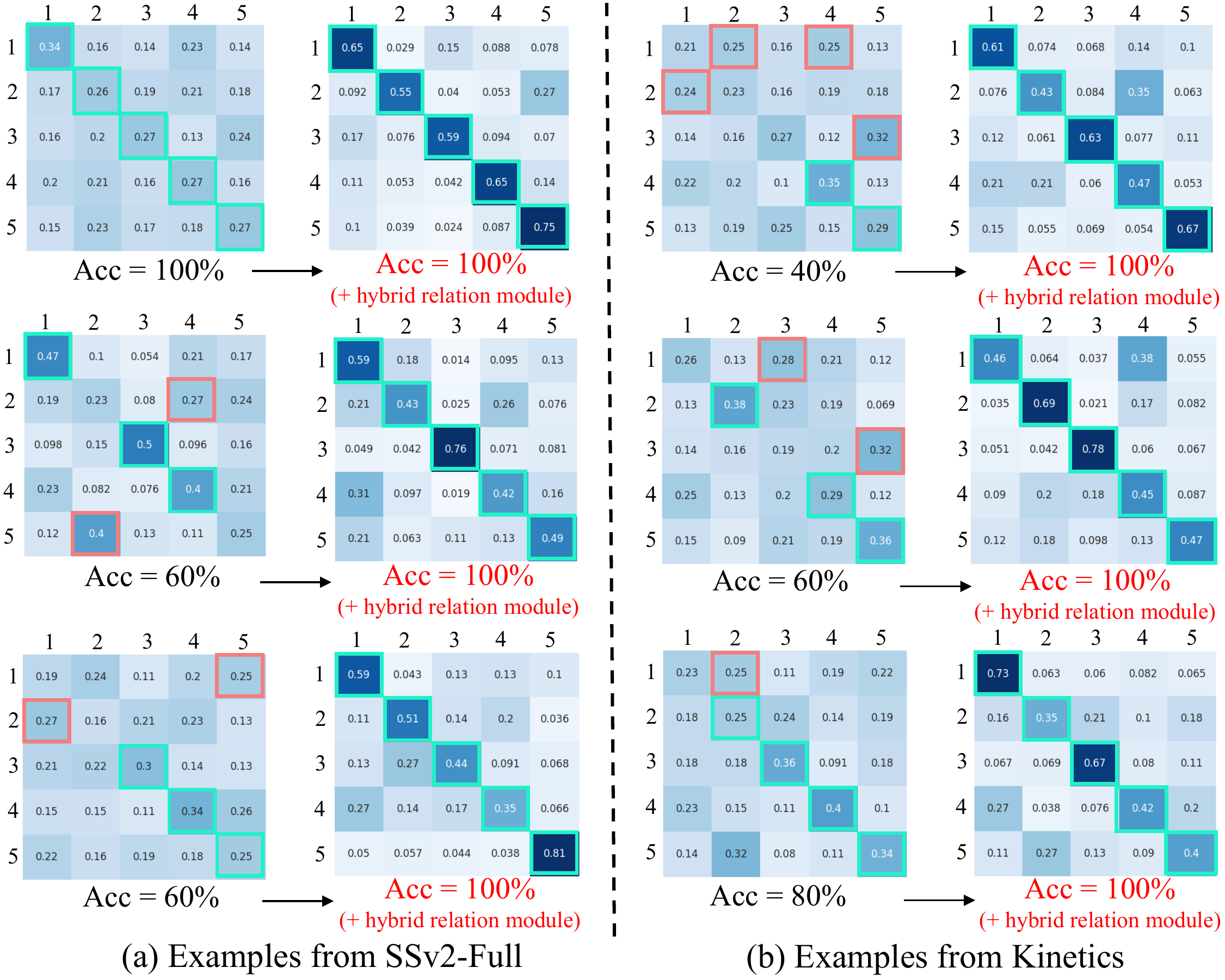}
\vspace{-3mm}
\caption{Similarity visualization of how query videos (rows) match to support videos (columns). 
The boxes of different colors correspond to: {\color[RGB]{33,241,201}correct match} and {\color[RGB]{242,126,126}incorrect match}.
}
\label{fig:visual_file_relation}
\vspace{-4mm}
\end{figure}
%
%
%

\subsection{Comparison with other matching approaches}
%
Our proposed set matching metric Bi-MHM aims to accurately find the corresponding video frames between video pairs by relaxing the strict temporal ordering constraints.
The following comparative experiments in Table~\ref{tab:matching-alignment} are carried out under the identical experimental setups, \ie, replace the OTAM directly with our Bi-MHM while keeping other settings unchanged. 
Results show that our Bi-MHM performs well and outperforms other temporal alignment methods (\eg, OTAM).
%
%
We further analyze different set matching approaches in Table~\ref{tab:matching}, and the results indicate Hausdorff distance is susceptible to noise interference, resulting in the mismatch and relatively poor performance. 
However, our Bi-MHM shows stability to noise and obtains better performance.
Furthermore, compared with the single directional metric,  our proposed bidirectional metric is more comprehensive to reflect the actual distances between videos and achieves better performance on few-shot tasks.
%
%

\subsection{Visualization results }
\begin{figure}[t]  
\centering
\includegraphics[width=0.44\textwidth]{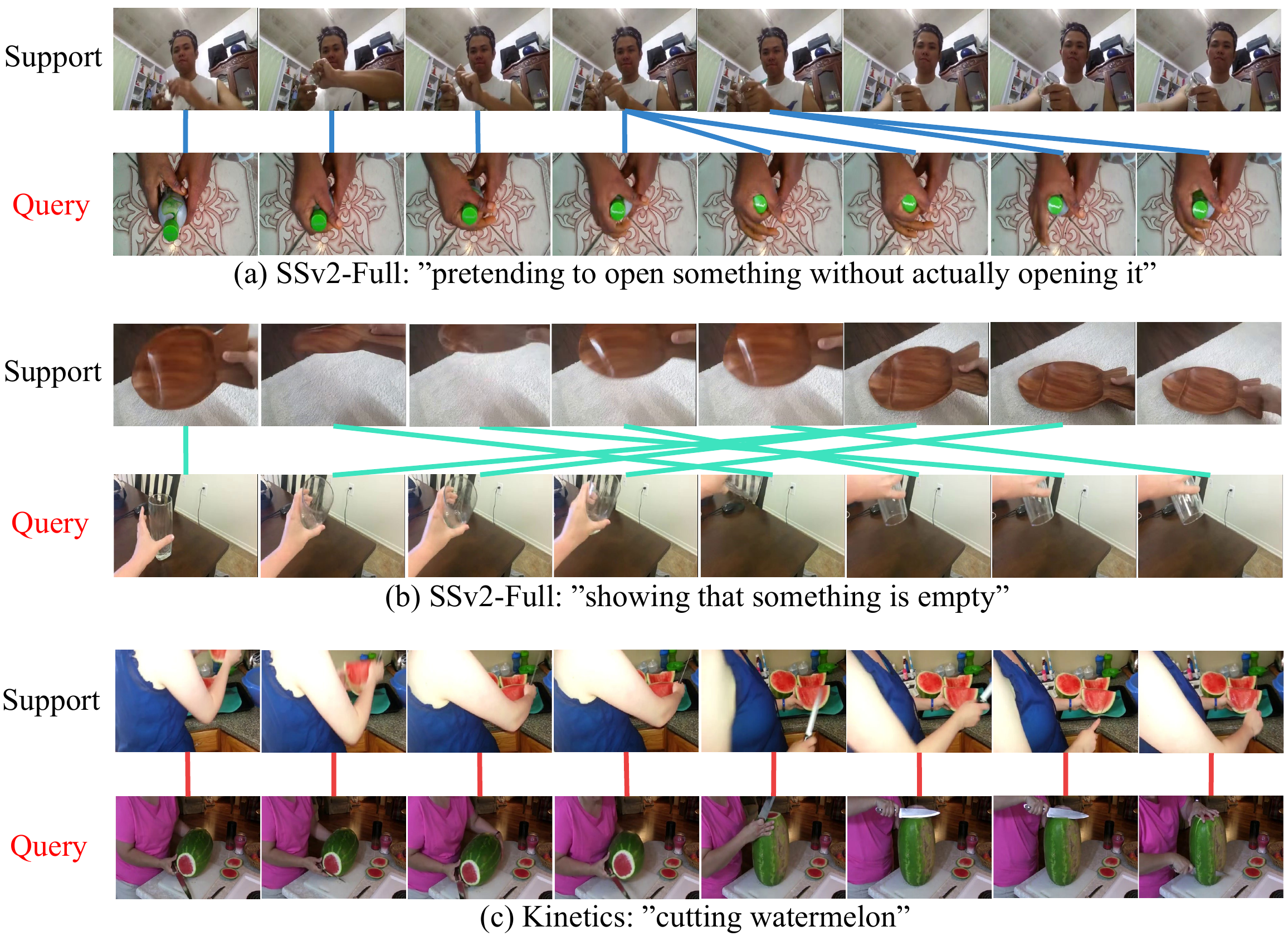}
\vspace{-3mm}
\caption{Visualization of matching results with the proposed set matching metric on SSv2-Full and Kinetics.
}
\label{fig:set_matching_visual}
\vspace{-3mm}
\end{figure}

%
%
To qualitatively show the discriminative capability of the learned task-specific features in our proposed method, we visualize the similarities between query and support videos with and without the hybrid relation module.
As depicted in Figure~\ref{fig:visual_file_relation}, by adding the hybrid relation module, the discrimination of features is significantly improved, contributing to predicting more accurately.
%
Additionally, the matching results of the set matching metric are visualized in Figure~\ref{fig:set_matching_visual}, and we can observe that our Bi-MHM is considerably flexible in dealing with alignment and misalignment.

%
%
%
%
%
\begin{table}
\centering
\small
\tablestyle{6pt}{1.1}
\begin{tabular}{lc|cc|cc}
\hspace{-1mm} Method & Backbone & Param &  FLOPs & Latency  &  Acc \\
\shline
\hspace{-1mm}  HyRSM & ResNet-18 & 13.8M & 3.64G & 36.5ms  & 46.6 \\
\hspace{-1mm}  HyRSM & ResNet-34 & 23.9M & 7.34G & 67.5ms   & 50.0 \\

\shline
\hspace{-1mm}  OTAM~\cite{OTAM} & ResNet-50  & \textbf{23.5M}   & \textbf{8.17G}  & 116.6ms & 42.8 \\
\hspace{-1mm}  TRX~\cite{TRX} &  ResNet-50  & 47.1M & 8.22G  & 94.6ms & 42.0 \\
\hspace{-1mm}  HyRSM & ResNet-50 & 65.6M & 8.36G & \textbf{83.5ms}  & \textbf{54.3} \\

\end{tabular}
\vspace{-3mm}
\caption{Complexity analysis for 5-way 1-shot SSv2-Full evaluation. The experiments are carried out on one Nvidia V100 GPU.
}
\label{tab:flops}
\vspace{-4mm}
\end{table}

%
%
\subsection{Limitations}
In order to further understand HyRSM, Table~\ref{tab:flops} illustrates its differences with OTAM and TRX in terms of parameters, computation, and runtime.
%
%
Notably, HyRSM introduces extra parameters (\ie, hybrid relation module), resulting in increased GPU memory and computational consumption.
Nevertheless, without complex non-parallel classifier heads, the whole inference speed of HyRSM is faster than OTAM and TRX.
We will further investigate how to reduce complexity with no loss of performance in the future.

\section{Conclusion}
In this work, we have proposed a hybrid relation guided set matching (HyRSM) approach for few-shot action recognition.
%
Firstly, we design a hybrid relation module to model the rich semantic relevance within one video and cross different videos in an episodic task to generate task-specific features.
%
%
Secondly, built upon the representative task-specific features, an efficient set matching metric is proposed to be resilient to misalignment and match videos accurately.
Experimental results demonstrate that our HyRSM achieves the state-of-the-art performance on the six standard benchmarks, including Kinetics, SSv2-Full, SSv2-Small, HMDB51, UCF101, and Epic-kitchens.

\section*{Acknowledgment}
%
This work is supported by the National Natural Science Foundation of China under grant 61871435, Fundamental Research Funds for the Central Universities no.2019kfyXKJC024, 111 Project on Computational Intelligence and Intelligent Control under Grant B18024, and Alibaba Group through Alibaba Research Intern Program.

%

{\small
\bibliographystyle{ieee_fullname}
\bibliography{egbib}
}

\clearpage
\appendix
\section*{Supplementary materials}
\section{Splits of Epic-kitchens}
Epic-kitchens~\cite{EPIC-100-2} is a large-scale first-view dataset and contains diverse unedited object interactions in kitchens.
In our experiment, we divide the dataset according to the verbs of the actions.

Meta-training set: 'take', 'put-down', 'open', 'turn-off', 'dry', 'hand', 'tie', 'remove', 'cut', 'pull-down', 'shake', 'drink', 'move', 'lift', 'stir', 'adjust', 'crush', 'taste', 'check', 'drain',  'sprinkle', 'empty', 'knead', 'spread-in', 'scoop', 'add', 'push', 'set-off', 'wear', 'fill', 'turn-down', 'measure', 'scrape', 'read', 'peel', 'smell', 'plug-in', 'flip', 'turn', 'enter', 'unscrew', 'screw-in', 'tap-on', 'break', 'fry', 'brush', 'scrub', 'spill', 'separate', 'immerse', 'rub-on', 'lower', 'stretch', 'slide', 'use', 'form-into', 'oil', 'sharpen', 'touch', 'let'.

Meta-testing set: 'wash', 'squeeze', 'turn-on', 'throw-in', 'close', 'put-into', 'fold', 'unfold', 'pour', 'tear', 'look-for', 'hold', 'roll', 'arrange', 'spray',  'wait', 'collect', 'turn-up', 'grate', 'wet'.

Note that there is no overlap between the meta-training set and the meta-testing set.
%
%
%


%
%
%
%
\begin{table}[h]
\centering
\small
\tablestyle{6pt}{1.1}
\begin{tabular}{l|c|cc}
\hspace{-1mm} Setting & Dataset &   1-shot  &  5-shot \\

\shline
\hspace{-1mm}  Support-only &  \multirow{2}{*}{SSv2-Full}    & 52.1  & 67.2 \\
\hspace{-1mm}  \textbf{Support\&Query (ours)} &  & \textbf{54.3}  & \textbf{69.0} \\
\shline
\hspace{-1mm}  Support-only & \multirow{2}{*}{Kinetics}   & 73.4  & 85.5 \\
\hspace{-1mm}  \textbf{Support\&Query (ours)} & & \textbf{73.7}  & \textbf{86.1} \\

\end{tabular}
\caption{Performance comparison with different relation modeling paradigms on SSv2-Full and Kinetics.
}
\label{tab:supportset&task}
\vspace{-3mm}
\end{table}
%
%
%
\begin{figure}[h]  
\centering
\includegraphics[width=0.45\textwidth]{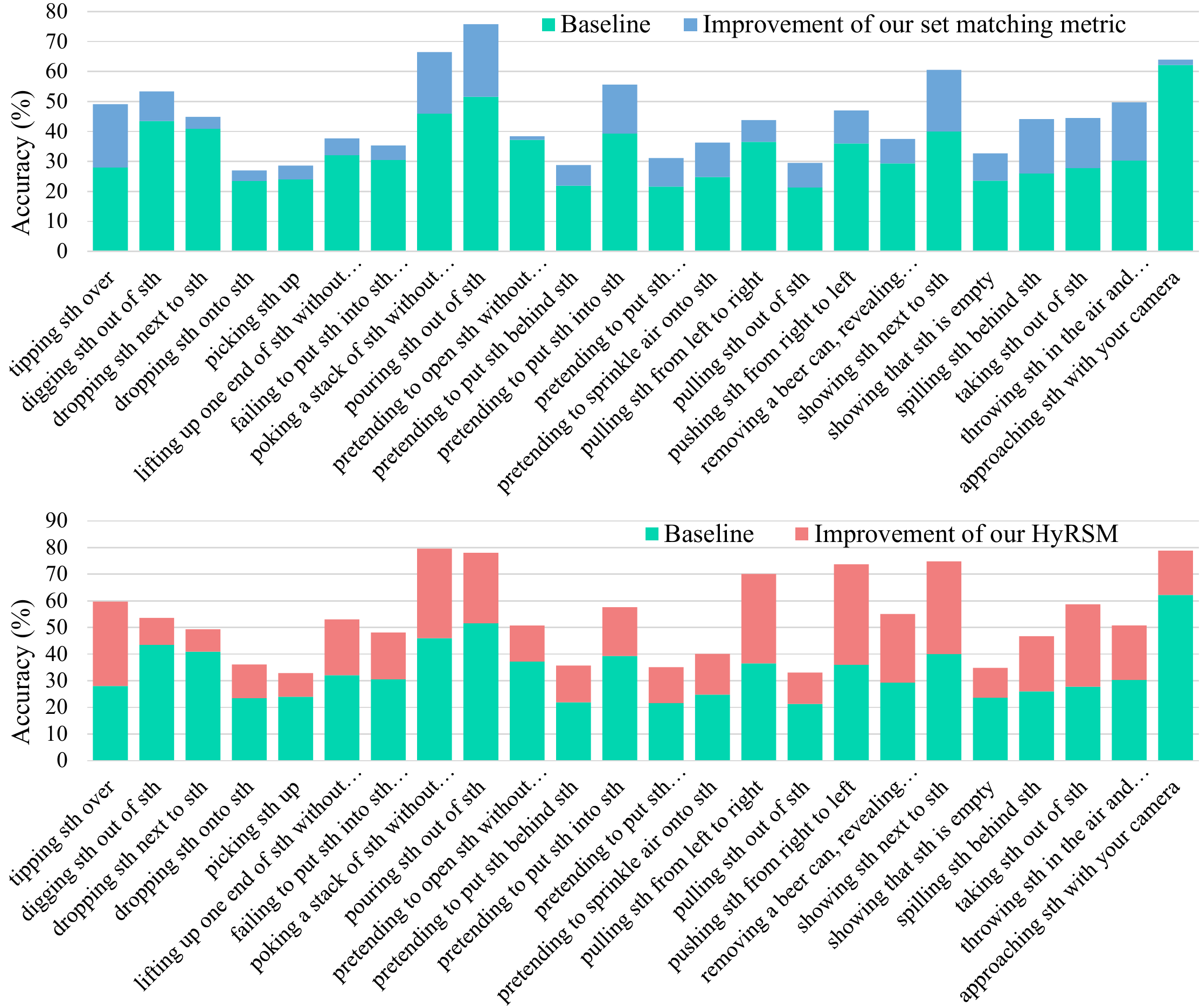}
\vspace{-3mm}
\caption{Category gain on the SSv2-Full dataset.}
\label{fig:Class_improvements}
\vspace{-3mm}
\end{figure}
%
%
%
%
\begin{figure}[t]  
\centering
\includegraphics[width=0.45\textwidth]{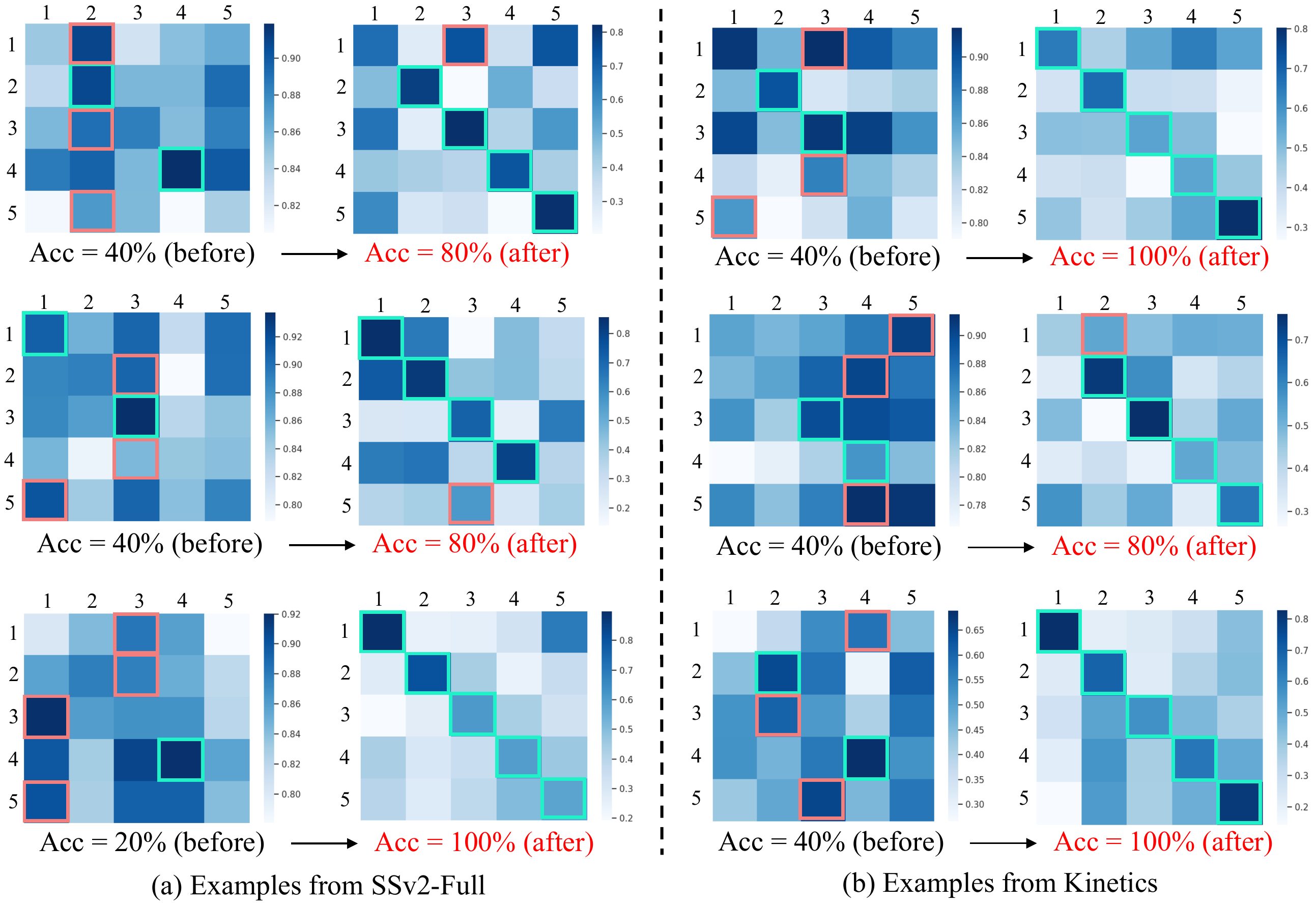}
\vspace{-3mm}
\caption{Similarity visualization of how query
videos (rows) match to support videos (columns) before and after the hybrid relation module in HyRSM. 
The boxes of different colors correspond to: {\color[RGB]{33,241,201}correct match} and {\color[RGB]{242,126,126}incorrect match}.
}
\vspace{-3mm}
\label{fig:visual_relation_before&after}
\end{figure}
%
%
%
%
\begin{figure*}[htbp]  
\centering
\includegraphics[width=0.98\textwidth]{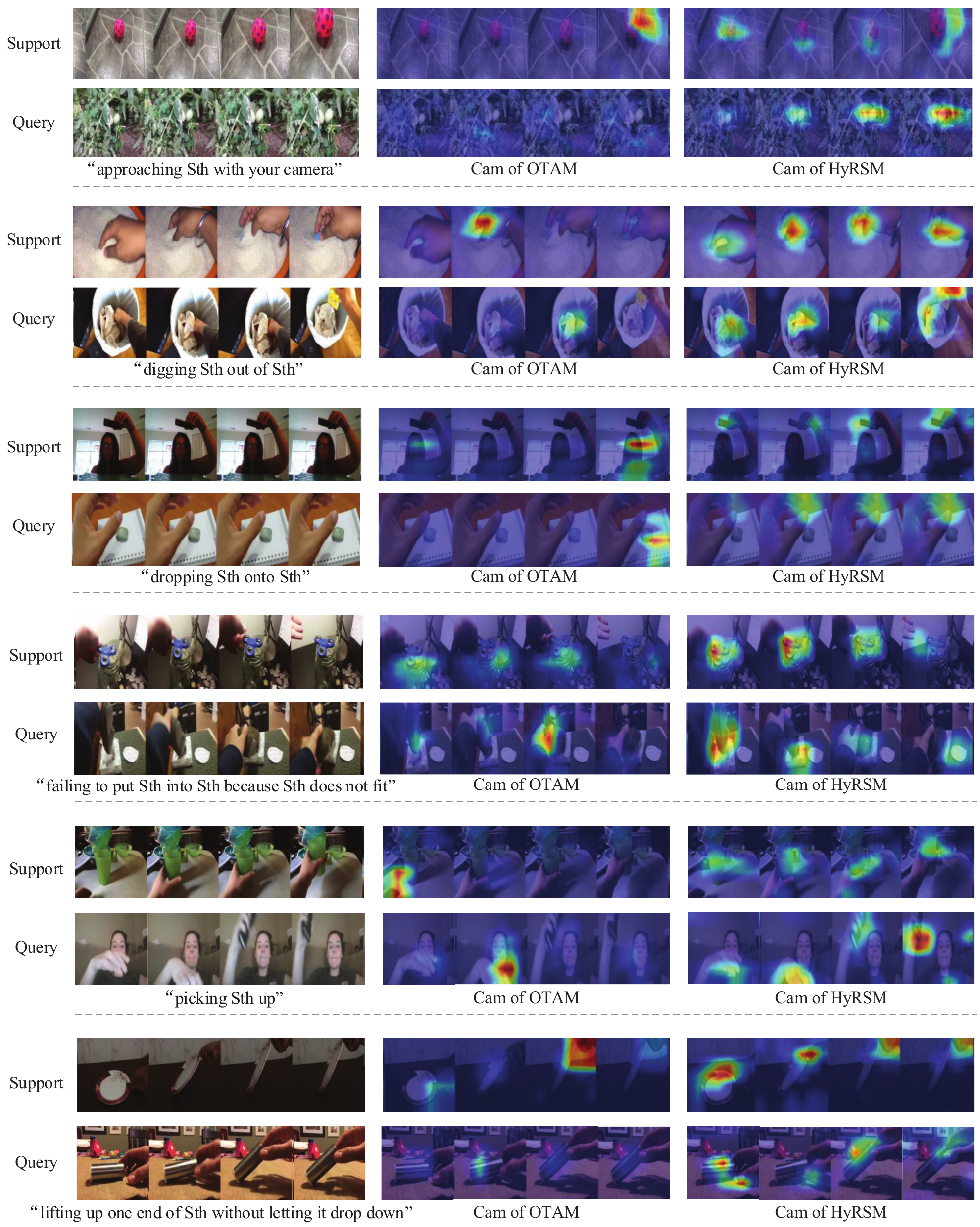}
\caption{Visualization of class activation maps (Cam) with Grad-CAM~\cite{grad-CAM} on SSv2-Full.
Corresponding to: original RGB images (left), Cam of OTAM~\cite{OTAM} (middle) and Cam of HyRSM (right).
}
\label{fig:Kinetics-visual}
\end{figure*}
\begin{figure*}[htbp]  
\centering
\includegraphics[width=0.97\textwidth]{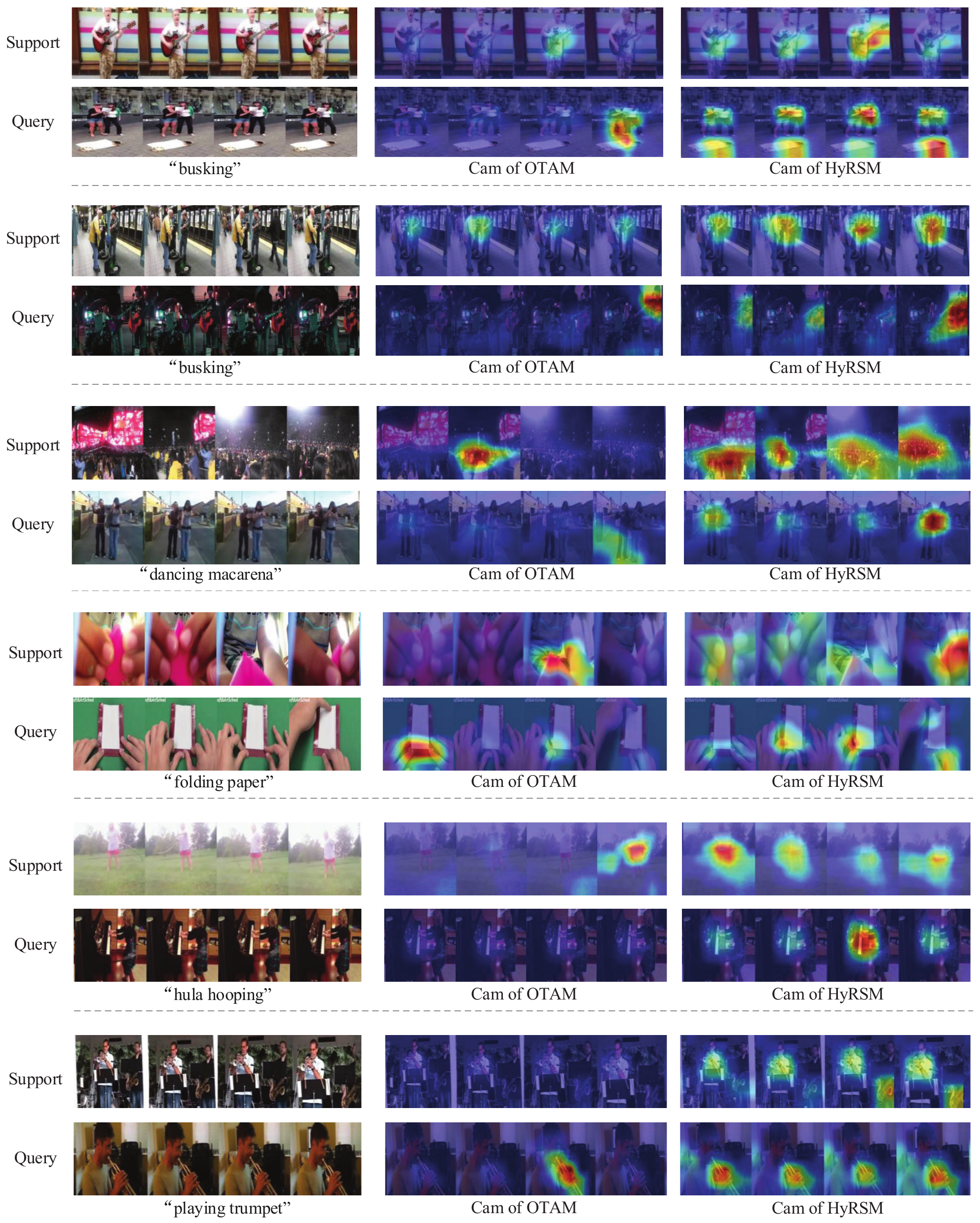}
\caption{ Visualization of class activation maps (Cam) with Grad-CAM~\cite{grad-CAM} on Kinetics.
Corresponding to: original RGB images (left), Cam of OTAM~\cite{OTAM} (middle) and Cam of HyRSM (right).
}
\label{fig:SSv2-visual}
\end{figure*}
\section{Other relation modeling forms}
Previous few-shot image classification methods of learning task-specific features have also achieved promising results~\cite{TapNet,Finding_task-relevant}. 
However, many of them use some complex and fixed operations to learn the dependencies between images, while our method is greatly simple and flexible. 
Moreover, most previous works only use the information within the support set to learn task-specific features, ignoring the correlation with query samples.
In our hybrid relation module, we add the query video to the pool of inter-relation modeling to extract relevant information suitable for query classification.
As illustrated in Table~\ref{tab:supportset&task}, we try to remove the query video from the pool, \ie, \emph{Support-only}, but we can observe that after removing the query video, the performance of 1-shot and 5-shot on SSv2-Full reduces by 2.2\% and 1.8\%, respectively.
There are similar conclusions on the Kinetics dataset.
This evidences that the proposed hybrid relation module is reasonable and can effectively extract task-related features, thereby promoting query classification performance.
%
%
%
\section{Class improvement}
In order to further analyze the performance improvement of each action category,
we compare the improvement of the proposed set matching metric and HyRSM compared to the baseline on SSv2-Full, as depicted in Figure~\ref{fig:Class_improvements}.
For the set matching metric, some action classes have limited improvements, \eg, "drop something onto something" and "pretending to open something without actually opening it", whereas some action classes have more than 20\% improvement, \eg, "tipping something over" and "showing something next to something".
For our HyRSM, the improvement of each category is more evident than the set matching metric.
In particular, "pulling something from left to right" and "pushing something from right to left" do not have significant increases in set matching metric but increase by more than 25\% in HyRSM.
This suggests that the hybrid relation module and the proposed set matching metric are strongly complementary.
%


\section{Visualization analysis}
To further demonstrate the effectiveness of our proposed hybrid relation module, we visualize the similarity maps of features before and after the hybrid relation module in HyRSM in Figure~\ref{fig:visual_relation_before&after}.
The results indicate that the features are improved significantly after refining by the hybrid relation module.
In addition, to qualitatively evaluate the proposed HyRSM, we compare the class activation maps visualization results of HyRSM to the competitive OTAM~\cite{OTAM}. 
As shown in Figure~\ref{fig:Kinetics-visual} and Figure~\ref{fig:SSv2-visual}, the features of OTAM usually contain non-target objects since it lacks the mechanism of learning task-specific embeddings for feature adaptation.
In contrast, our proposed HyRSM processes the query and support videos with adaptive relation modeling operation, which allows it to focus on the different target
objects.

\section{Relation modeling operations}
In the literature~\cite{Transformer,BILSTM,LSTM,wang2021proposal,wang2021weakly,GRU,non-local}, there are many alternative relation modeling operations, including multi-head self-attention (MSA), Transformer, Bi-LSTM, Bi-GRU, \etc.

\vspace{+2pt}
\noindent \textbf{Multi-head self-attention} 
mechanism operates on the triple query $Q$, key $K$ and value $V$, and relies on scaled dot-product attention operator:
\begin{equation}
    Attention(Q; K; V ) = softmax(\frac{QK^{T}}{\sqrt{d_{k} } } )V
\end{equation}
where $d_{k}$ a scaling factor equal to the channel dimension of key $K$.
Multi-head self-attention obtains $h$ different heads and each head computes scaled dot-product attention representations of triple $(Q, K, V)$, concatenates the intermediates, and projects the concatenation through a fully connected layer.
The formula can be expressed as:
\begin{align}
   head_{i} &= Attention(QW_{i}^{q}; KW_{i}^{k}; VW_{i}^{v})\\
   MSA(Q; K; V ) &= concat_{i}(head_i)W^{\prime}, 1\le i \le h.
\end{align}
where the $W_{i}^{q}$, $W_{i}^{k}$, $W_{i}^{v}$ and $W^{\prime}$ are fully connected layer parameters.
Finally, a residual connection operation is employed to generate the final aggregated representation:
\begin{equation}
    f_{msa} = MSA(f; f; f) + f
\end{equation}
where $f$ comes from the output of the previous layer. Note that query, key and value are the same in self-attention. 
\vspace{+2pt}
\noindent \textbf{Transformer}
is a state-of-the-art architecture for natural language processing~\cite{Transformer,Bert,transformer-xl}.
Recently, it has been widely used in the field of compute vision~\cite{DETR,DETR-Seg,ViT,OadTR} due to its excellent contextual modeling ability, and has achieved significant performances.
Transformer contains two sub-layers: (a) a multi-head self-attention
layer (MSA), and (b) a feed-forward network (FFN).
Formulaic expression is:
\begin{equation}
    f_{transformer} = FFN(f_{msa}) + f_{msa}
\end{equation}
where FFN contains two MLP layers with a GELU non-linearity~\cite{GELU}.

\vspace{+2pt}
\noindent \textbf{Bi-LSTM}
is an bidirectional extension of the Long Short-Term Memory (LSTM) with the ability of managing variable-length sequence inputs.
Generally, an LSTM consists of three gates: forget gate, input gate and output gate. 
The forget gate controls what the existing information needs to be preserved/removed from the memory. 
The input gate makes the decision of whether the new arrival will be added.
The output gate uses a sigmoid layer to determine which part of memory attributes to the final output.
The mathematical equations are:
\begin{align}
   f_{t} &= \sigma(W_{f_h}[h_{t-1}] + W_{f_{x}}[x_{t}] + b_{f})\\
   i_{t} &= \sigma(W_{i_h}[h_{t-1}]+ W_{i_{x}}[x_{t}]+ b_{i})\\
   \widetilde{c_{t}} &= \tanh(W_{c_h}[h_{t-1}]+ W_{c_{x}}[x_{t}]+ b_{c})\\
   c_{t} &= f_{t}\ast c_{t-1} + i_{t} \ast \widetilde{c_{t}} \\
   o_{t} &= \sigma(W_{o_h}[h_{t-1}]+W_{o_{x}}[x_{t}]+ b_{o})\\
   h_{t} &= o_{t} \ast tanh(c_{t})
\end{align}
where $f_{t}$ is the value of the forget gate, $o_{t}$ is the output result, and $h_{t}$ is the output memory.
In Bi-LSTM, two LSTMs are applied to the input and the given input data is utilized twice for training (\ie, first from left to right, and then from right to left).
Thus, Bi-LSTM can be used for sequence data to learn long-term temporal dependencies.

\vspace{+2pt}
\noindent \textbf{Bi-GRU}
is a variant of Gated Recurrent Unit (GRU) and have been shown to perform well with long sequence applications~\cite{Bigru_use_1,Bigru_use_2}.
In general, the GRU cell contains two gates: update gate and reset gate.
The update gate $z_{t}$ determines how much information is retained in the previous hidden state and how much new information is added to the memory.
The reset gate $r_{t}$ controls how much past information needs to be forgotten.
The formula can be expressed as:
\begin{align}
   z_{t} &= \sigma(W_{z}[x_{t}] + U_{z}[h_{t-1}] + b_{z}) \\
   r_{t} &= \sigma(W_{r}[x_{t}] + U_{r}[h_{t-1}] + b_{r}) \\
  \widetilde{h_{t}} &= g(W_{h}[x_{t}]+U_{h}[(r_{t}\ast h_{t-1})] + b_{h})  \\
  h_{t} &= (1-z_{t})\ast h_{t-1} + z_{t} \ast \widetilde{h_{t}}
\end{align}
where $x_{t}$ is the current input and $ h_{t} $ is the output hidden state.

\end{document}